\documentclass[10pt]{article}

\usepackage{geometry}
\usepackage[backend=biber,sorting=none,citestyle=numeric-comp,bibstyle=ieee]{biblatex}
\DeclareFieldFormat{url}{Available\addcolon\space\url{#1}}

\usepackage[hidelinks]{hyperref}
\hypersetup{colorlinks=true, allcolors=blue}

\usepackage{graphicx}
\usepackage{amsmath}
\usepackage{amsfonts}
\usepackage{amssymb}
\usepackage[mathscr]{eucal}
\usepackage{bm}
\usepackage[normalem]{ulem}
\usepackage{url}
\usepackage{tablefootnote}
\usepackage{placeins}
\usepackage{bbm}
\usepackage{mathtools} 
\usepackage[title]{appendix}

\usepackage{tikz}
\usepackage{etoolbox} 
\usepackage{listofitems} 

\tikzset{>=latex} 
\colorlet{myred}{red!80!black}
\colorlet{myblue}{blue!80!black}
\colorlet{mygreen}{green!60!black}
\colorlet{mydarkred}{myred!40!black}
\colorlet{mydarkblue}{myblue!40!black}
\colorlet{mydarkgreen}{mygreen!40!black}
\tikzstyle{node}=[very thick,circle,draw=myblue,minimum size=22,inner sep=0.5,outer sep=0.6]
\tikzstyle{connect}=[->,thick,mydarkblue,shorten >=1]
\tikzset{ 
  node 1/.style={node,mydarkgreen,draw=mygreen,fill=mygreen!25},
  node 2/.style={node,mydarkblue,draw=myblue,fill=myblue!20},
  node 3/.style={node,mydarkred,draw=myred,fill=myred!20},
}
\def\nstyle{int(\lay<\Nnodlen?min(2,\lay):3)} 

\usepackage{xcolor}

\newcommand{\dsphere}{\mathbb{S}^{d-1}}
\newcommand{\RR}{\mathbb{R}}

\usepackage{authblk}

\title{Deep learning joint extremes of metocean variables using the SPAR model}
\author[1]{E. Mackay\thanks{email: e.mackay@exeter.ac.uk,\quad ORCID: 0000-0001-7121-4231}}
\author[2,3]{C.J.R. Murphy-Barltrop}
\author[4]{J. Richards}
\author[5]{P. Jonathan}
\affil[1]{University of Exeter, UK}
\affil[2]{Technische Universität Dresden, Germany}
\affil[3]{ScaDS.AI, Germany}
\affil[4]{School of Mathematics and Maxwell Institute for Mathematical Sciences, University of Edinburgh, UK}
\affil[5]{Lancaster University, UK}
\date{\today}

\begin{document}

\maketitle

\begin{abstract}
This paper presents a novel deep learning framework for estimating multivariate joint extremes of metocean variables, based on the Semi-Parametric Angular-Radial (SPAR) model. When considered in polar coordinates, the problem of modelling multivariate extremes is transformed to one of modelling an angular density, and the tail of a univariate radial variable conditioned on angle. In the SPAR approach, the tail of the radial variable is modelled using a generalised Pareto (GP) distribution, providing a natural extension of univariate extreme value theory to the multivariate setting. In this work, we show how the method can be applied in higher dimensions, using a case study for five metocean variables: wind speed, wind direction, wave height, wave period, and wave direction. The angular variable is modelled using a kernel density method, while the parameters of the GP model are approximated using fully-connected deep neural networks. Our approach provides great flexibility in the dependence structures that can be represented, together with computationally efficient routines for training the model. Furthermore, the application of the method requires fewer assumptions about the underlying distribution(s) compared to existing approaches, and an asymptotically justified means for extrapolating outside the range of observations. Using various diagnostic plots, we show that the fitted models provide a good description of the joint extremes of the metocean variables considered.
\end{abstract}

\section{Introduction} \label{sec:intro}
Many problems in offshore and coastal engineering require estimation of joint extremes for metocean variables. Responses of offshore and coastal structures are dependent on multiple variables, such as wind speed and direction, wave height, period and direction, flow speed and direction. Providing reliable estimates of the joint extremes in this setting is a challenging problem for metocean engineers. Various design standards recommend the use of the environmental contour method \cite{haver2008environmental}. Some types of contour can be estimated without an explicit model for the joint distribution of variables \cite{Derbanne2019, mackay2023-DIFORM}. However, environmental contour methods typically make simplifying assumptions and only give approximate estimates of long-term extreme responses \cite{haselsteiner2022_NREL, speers2024}. Full probabilistic analysis of long-term extreme responses requires a model for the joint density of the relevant metocean variables. A wide range of approaches have been proposed for estimating joint densities. In the offshore engineering literature, the two most popular approaches are global hierarchical models and copula models -- see, e.g., \cite{Ross2020, haselsteiner2021_benchmark}. 

Let $\mathbf{X} = (X_1,...,X_d) \in \mathbb{R}^d$, $d\ge 2$ denote a continuous random vector with joint density function $f_{\mathbf{X}}$, and marginal density and cumulative distribution functions $f_{X_j}$ and $F_{X_j},$ respectively, for $j=1,...,d$. In the global hierarchical approach \cite{bitner2015joint, horn2018new, cheng2019long, simao2022multi}, the joint density is written as
\begin{equation}
    f_{\mathbf{X}}(\mathbf{x}) = f_{X_1}(x_1) \, f_{X_2 | X_1}(x_2 | x_1) \cdots f_{X_d | (X_1,...,X_{d-1})}(x_d | x_1,...,x_{d-1}),
\end{equation}
where $f_{X_j | (X_1,...,X_{j-1})}$ is the density of $X_j$ conditional on $(X_1,...,X_{j-1})$ for $j \in \{2,\hdots,d\}$. Inference typically involves selecting parametric forms for $f_{X_1}$, $f_{X_2 | X_1}$, ..., $f_{X_d | (X_1,...,X_{d-1})}$ and estimating relations between the parameters of the conditional densities and the conditioning variables. There are various problems with this approach. Firstly, there is no a priori reason to suppose that variables follow any particular parametric distribution, and misspecified models can have dramatic consequences when approximating dependence structures. Secondly, a model fit to all of the observations does not guarantee a good fit to the tail, which is the region of interest for extremes. Finally, the models for the parameters of the conditional densities are usually based on ad hoc assumptions, and provide no rationale for extrapolating outside the range of observations. In many cases, it has been shown that such models provide a poor fit to observed data, especially in extreme regions \cite{deHauteclocque2022}. 

For copula modelling, the joint density is written as 
\begin{equation}
    f_{\mathbf{X}}(\mathbf{x}) = f_{X_1}(x_1) \cdots f_{X_d}(x_d) c(F_{X_1}(x_1), \cdots, F_{X_d}(x_d)),
\end{equation}
where $c:[0,1]^d\mapsto[0,\infty)$ is the copula density of $\mathbf{X}$ \cite{Joe2015}. In this case, inference involves choosing parametric models for both the marginal densities and for $c$. As with the global hierarchical approach, there are no a priori reasons to choose particular models. Similarly, fitting to all observations does not guarantee a good fit to the tails. Moreover, different copula models have very different behaviours in the joint tail regions, meaning extrapolation can vary substantially for different choices of copula model \cite{Joe2015}.

There are also a wide range of methods in the statistical literature for modelling joint extremes (e.g., \cite{Beirlant2004, Ledford1997, Wadsworth2013}). However, many of these approaches make strong assumptions about the dependence structure, or copula, which are often not supported by environmental datasets \cite{huser2024modeling}. The most popular choice for metocean variables is the conditional extremes model \cite{Heffernan2004}, which describes the joint distribution of variables conditional on at least one variable being large. The key limitation of this approach is that it only characterises the region of variable space where the conditioning variable is large, and inferences made using different conditioning variables are not necessarily consistent \cite{Liu2014}. A further limitation of this method (and other methods in the multivariate extremes literature) is that it requires a transformation of the margins to a standard scale. This requires first estimating the marginal distributions for each variable -- a process which is subject to uncertainty. Furthermore, it has been demonstrated that poor marginal estimates greatly affect the quality of the resulting multivariate inference \cite{towe2023estimation}.

In this paper, we discuss the application of a new method, introduced in \cite{mackay2022imex}, which overcomes the limitations of existing approaches and provides a general, flexible framework for modelling multivariate extremes. The model is referred to as the Semi-Parametric Angular-Radial (SPAR) model. The SPAR model provides a framework for estimating multivariate extremes that does not require strong assumptions about the form of the margins or dependence structure, and provides a justified means of extrapolating outside the range of observations. Moreover, the model is only fitted to extreme observations, meaning that no assumptions are required about the bulk of the distribution. Theoretical aspects of the SPAR model are presented in \cite{mackay2023-SPAR}, and an inference approach in a two-dimensional setting is provided in \cite{murphy2024inference, mackay2025spar}. The purpose of this paper is to extend the modelling method to the general multivariate setting, with two or more dimensions. The two-dimensional inference scheme in \cite{murphy2024inference, mackay2025spar} utilised cubic regression splines to model the angular dependence of the radial distribution parameters. In this work we take a different approach, and adopt a deep learning scheme, where the radial distribution parameters are modelled using artificial neural networks. As discussed further in Section \ref{sec:inference}, this offers some computational advantages in higher-dimensional settings.  

The paper is organised as follows. Section \ref{sec:theory} describes a brief overview of the theoretical aspects of the model. Our deep learning approach for estimating the radial component of the SPAR model is introduced in Section~\ref{sec:inference}. Section~\ref{sec:application} presents an example application of the model to a five-dimensional problem: estimating the joint extremes of wind speed, wave direction, wave height, wave period, and wave direction. We discuss the challenges that arise for these particular variables, and how well the model assumptions are satisfied in this setting. Several novel types of diagnostic plots are introduced to assess the fit of the model. A discussion and outlook on future work is presented in Section~\ref{sec:discussion}. Finally, 
conclusions are presented Section~\ref{sec:conclusions} .

\section{Theory}
\label{sec:theory} 
\subsection{SPAR model definition} \label{sec:SPAR_def}
The SPAR model can be viewed as an extension of the univariate peaks-over-threshold (POT) method to the multivariate setting. It involves a transformation of variables to angular-radial coordinates, and then models the upper-tail of the radial variable, conditional on angle, using a non-stationary generalised Pareto (GP) model. Suppose that we have a continuous random vector $\mathbf{X} = (X_1,...,X_d) \in \mathbb{R}^d$ with joint density function $f_{\mathbf{X}}$. We define radial and angular variables as
\begin{equation}
    R = \|\mathbf{X}\|_2, \quad \mathbf{W} = \mathbf{X} / R,
\end{equation}
where $\|\cdot\|_2$ is the $L^2$ or Euclidean norm, defined by $\|(x_1,...,x_d)\|_2 = (x_1^2+ \cdots + x_d^2)^{1/2}$. Note that $R\in [0,\infty)$ and $\mathbf{W}\in \mathbb{S}^{d-1}$, where $\mathbb{S}^{d-1} = \{\mathbf{x}\in\mathbb{R}^d : \|\mathbf{x}\|_2 = 1\}$ is the unit hypersphere in $\mathbb{R}^d$. The joint density function of $(R,\mathbf{W})$ is related to $f_{\mathbf{X}}$ via 
\begin{equation} \label{eq:frw}
    f_{R,\mathbf{W}}(r,\mathbf{w}) = r^{d-1} \, f_{\mathbf{X}}(r \mathbf{w}), \quad r\ge0,\, \mathbf{w}\in\dsphere,
\end{equation}
where $r^{d-1}$ is the Jacobian determinant for the transformation $\mathbf{X}\rightarrow (R,\mathbf{W})$. As for global hierarchical models, the angular-radial joint density can be written in conditional form as:
\begin{equation} \label{eq:frw_cond}
    f_{R,\mathbf{W}}(r,\mathbf{w}) = f_{\mathbf{W}}(\mathbf{w}) \,f_{R|\mathbf{W}}(r|\mathbf{w}).
\end{equation}
Noting that $\mathbf{X}=R\mathbf{W}$, and that $\mathbf{W}$ lies on the surface of the unit hypersphere, we can see that the `extreme' parts of the distribution of $\mathbf{X}$ correspond to large values of the radial variable at any given angle. Therefore, the problem of modelling multivariate extremes is transformed to that of modelling an angular density $f_{\mathbf{W}}$ and the tail of the conditional radial density $f_{R|\mathbf{W}}$. For a given angle $\mathbf{w}$, the density $f_{R|\mathbf{W}}(r|\mathbf{w})$ is univariate. Univariate extreme value theory suggests that a suitable model for the tail of $f_{R|\mathbf{W}}$ is the GP distribution, with parameters conditional on angle (e.g., \cite{Coles2001}). This motivates the SPAR model, whereby parametric and non-parametric models are used to model the conditional radial and angular distributions, respectively. Define a threshold function $u(\mathbf{w})>0$ to be the quantile of $R|(\mathbf{W} = \mathbf{w})$ with exceedance probability $\zeta\in(0,1)$, with $\zeta$ close to 0, i.e., the solution of $\zeta = \Pr (R>u(\mathbf{w}) | \mathbf{W} = \mathbf{w})$. The SPAR model can be written as
\begin{equation} \label{eq:frw_SPAR}
    f_{R,\mathbf{W}} (r,\mathbf{w}) = \zeta f_{\mathbf{W}} (\mathbf{w}) f_{\rm GP} (r - u(\mathbf{w}); \xi(\mathbf{w}), \sigma (\mathbf{w})), \quad r>u(\mathbf{w}),\, \mathbf{w}\in\dsphere,
\end{equation}
where $f_{\rm GP}$ is the GP density function, and $\xi(\mathbf{w}) \in \mathbb{R}$ and {$\sigma (\mathbf{w})>0$} are shape and scale parameters, respectively, given as functions of the angle $\mathbf{w}$. The GP density function is given by
\begin{align} \label{eqn:gp_dens}
    f_{\rm GP}(r;\xi,\sigma) = 
    \begin{cases}
    \dfrac{1}{\sigma} \left(1+\xi\dfrac{r}{\sigma}\right)^{-1-\tfrac{1}{\xi}}, & \xi\neq 0,\\[10pt]
    \dfrac{1}{\sigma} \exp\left(-\dfrac{r}{\sigma}\right), & \xi=0,
    \end{cases}
\end{align}
which is supported on $0\leq r \leq r^F$, where $r^F=\infty$ for $\xi\geq 0$ and {$r^F = -\sigma/\xi$} for $\xi<0$. 

Many non-parametric methods for estimation of densities, such as kernel density methods, mixture models or spline-based methods, assume that the density is finite and continuous. Similarly, many representations for non-stationary modelling of parametric distributions assume that the parameter functions are finite and continuous. Therefore, to simplify our inference, we assume that the angular density $f_{\mathbf{W}}$, threshold function $u(\mathbf{w})$, and GP parameter functions, $\xi(\mathbf{w})$ and $\sigma (\mathbf{w}),$ are finite and continuous with respect to the angle $\mathbf{w}$. 

After estimation of the angular density and GP parameter functions, \eqref{eq:frw} and \eqref{eq:frw_SPAR} can be combined to obtain the SPAR estimate of the joint density in the original variable space for observations satisfying $r>u(\mathbf{w})$, i.e., 
\begin{equation}
    f_{\mathbf{X}} (r\mathbf{w}) = \zeta \, r^{1-d} \, f_{\mathbf{W}} (\mathbf{w}) \, f_{\rm GP} (r-u(\mathbf{w}); \xi(\mathbf{w}), \sigma (\mathbf{w})).
\end{equation}
Calculating marginal and joint probabilities using the SPAR model then involves either integration of the joint density over specified angular and radial domains, or via Monte Carlo techniques, i.e., by simulating from the estimated model and deriving probability estimates empirically. To simulate from the SPAR model, we first draw an angle $\mathbf{w}$ from $f_{\mathbf{W}}$, then use inversion sampling to generate a corresponding value $z$ from the GP distribution with parameter vector $(\xi(\mathbf{w}), \sigma(\mathbf{w}))$, and finally define a radial value $r=u(\mathbf{w})+z$. The pair $(r,\mathbf{w})$ is then a random sample from the SPAR model. This can be converted back to the original variable space using the inverse transformation $\mathbf{x}=r\mathbf{w}$. As the SPAR model is only fitted to observations for which $r>u(\mathbf{w})$, one can create a sample (of the original random vector $\mathbf{X}$) of size $N$ by simulating $\zeta N$ points from the SPAR model, and then resampling $(1-\zeta) N$ points from observations with $r\leq u(\mathbf{w})$. The rationale for this is that there should be a sufficient number of observations within the body of the distribution to obtain a reasonable estimate from resampling. 

\subsection{Exceedance probability contours}
As described in \cite{mackay2025spar}, the SPAR model provides an explicit means for calculating a contour with a specified exceedance probability $\beta \in(0,\zeta]$. The radius of the contour at angle $\mathbf{w}$ is simply the quantile of the GP distribution at exceedance probability $\beta /\zeta$, given by
\begin{equation}\label{eq:SPAR_contour}
    r_\beta (\mathbf{w}) = u(\mathbf{w}) + \frac{\sigma(\mathbf{w})}{\xi(\mathbf{w})} \left(\left(\frac{\beta}{\zeta}\right)^{-\xi(\mathbf{w})} - 1\right).
\end{equation}
This contour is defined in terms of the probability of an observation falling anywhere outside the contour region, or the `total exceedance probability'. As such, these contours are more conservative than those defined in terms of marginal exceedance probabilities, such as IFORM contours (or variants thereof), with the conservatism increasing with the number of dimensions \cite{mackay2021marginal}. Moreover, if the primary interest of the analysis is to estimate environmental contours, then the use of the SPAR model is not necessary. Instead, we recommend the use of the Direct-IFORM method \cite{Derbanne2019, mackay2023-DIFORM}, which does not require a model for the joint density or any assumptions about the dependence structure between the variables. 

The contours defined in \eqref{eq:SPAR_contour} can be projected into two dimensions. In Cartesian space, points on the contour are given by $\mathbf{x} = r_\beta (\mathbf{w}) \, \mathbf{w}$ for $\mathbf{w}\in\dsphere$. The radii $r_\beta\ge 0$ can be computed for a discrete set of points on the sphere (see Appendix \ref{app:L1L2_projection}), and the projection into dimensions $i,j\in\{1,...,d\}$ is just the $i$ and $j$ components of $\mathbf{x}$. However, in our application it was found that these contours do not provide a useful diagnostic tool. This is because they do not account for the angular density. In regions of low angular density, estimates of the contours can be highly uncertain. Simpson and Tawn \cite{simpson2024inference} proposed to mitigate for this by either setting the radius to zero when the angular density is zero, or adjusting the exceedance level to account for the angular density. However, the latter approach requires an integral over angles that becomes impractical in higher dimensions. We have therefore opted to use alternative diagnostics, described in Section \ref{sec:radial_assess}.

\section{Inference} \label{sec:inference} 
Inference for the SPAR model involves estimating the angular density, and the GP threshold and parameter functions. Since the angular-radial density factorises (see \eqref{eq:frw_cond}), these problems are separable: inference for $f_{\mathbf{W}}$ can be conducted independently of that for $(u(\mathbf{w}), \sigma(\mathbf{w}), \xi(\mathbf{w}))$. Inference for the angular density is discussed in Section~\ref{sec:ang_dens}, and modelling of the conditional radial variable is discussed in Section~\ref{sec:rad_dens}. Code for fitting our model is available at \url{https://github.com/callumbarltrop/DeepSPAR}. 

\subsection{Angular modelling} \label{sec:ang_dens}
In the SPAR method we estimate the angular density of all observations -- see \eqref{eq:frw_SPAR}. This is different to `classical' multivariate extreme value modelling, where it is only the angular density of `extreme' observations that is modelled (see e.g. \cite{Coles1991}. Using the entire sample gives more observations to work with, reducing sampling uncertainties. However, in contrast classical multivariate extremes, in the SPAR approach there are no theoretically-motivated parametric models for the angular density, meaning that flexible modelling approaches are required. 

Estimation of densities on the hypersphere $\mathbb{S}^{d-1}$ is part of a discipline known as directional statistics \cite{mardia2000, ley2017modern}. The key difference from estimation of densities on $\mathbb{R}^d$ is that the surface of the hypersphere is periodic and bounded, and so distributions defined on $\mathbb{S}^{d-1}$ must conserve these constraints. Various parametric and non-parametric approaches have been developed for estimating densities on the hypersphere, which are directly analogous to approaches used in Euclidean space. These include kernel density (KD) estimation \cite{hall1987kernel, bai1989kernel}, mixture models \cite{mooney2003fitting, fu2008modified, hornik2014movmf}, and spline-based methods \cite{ferreira2008directional}. See \cite{Pewsey2021} for a recent review of non-parametric approaches, and \cite{wessel2025comparison} for a comparison of various generative deep learning methods for modelling angular densities.

In keeping with the previous uses of SPAR in two dimensions, in this work we use a KD method for the angular modelling. The advantages of KD estimation are that it is simple to implement and fast to sample from. The main downside is that explicit calculation of the density can be slow for large sample sizes. However, for our purpose, simulation from the model is more important than explicit calculation of the density, so this drawback is less significant. 

\subsubsection{Kernel density estimation for spherical data}
Suppose that $K(\mathbf{w}; \bm{\mu}, \kappa)$ is a density function on $\mathbb{S}^{d-1}$, with mean direction $\bm{\mu}\in\dsphere$, bandwidth parameter $\kappa\ge 0$ that controls the concentration of the density around $\bm{\mu}$. Then the KD estimate of the angular density $f_{\mathbf{W}}$ from a sample of $n$ observations $\{\mathbf{w}_i\}_{i=1,...,n} \subset \dsphere$ is given by
\begin{equation} \label{eq:KD}
	\hat{f}_{\mathbf{W}}(\mathbf{w}) = \frac{1}{n} \sum_{i=1}^n K(\mathbf{w}; \mathbf{w}_i, \kappa),
\end{equation}
That is, the estimate is effectively a mixture of $n$ kernels, centred at each data point. From a modelling perspective, there are two key choices to make: the type of kernel to use, and the determination of the bandwidth. 

\subsubsection{Choice of kernel}
The choice of kernel function (amongst commonly-used parametric forms for angular densities) has a relatively minor impact on the accuracy of the density estimate compared to the choice of bandwidth. While different kernels can slightly alter the shape of the density estimate, the bandwidth parameter, which controls the amount of smoothing, is the most crucial factor affecting the accuracy \cite{scott2015multivariate}. Here we are mainly guided by computational considerations. The most common choice of kernel for KD models on the sphere is the von Mises-Fisher (vMF) distribution \cite{fisher1953dispersion}. However, simulation from this distribution requires rejection sampling, which can be slow. Therefore, we opt to use the \textit{power spherical} (PS) distribution \cite{deCao2020psd}. The PS and vMF distributions are both rotationally symmetric distribution about the mean direction, and have similar shapes. However, simulation from the PS distribution is much faster, as discussed further in Appendix \ref{app:PSD_simulation}. For $\mathbf{w}\in\dsphere$ the density of the PS distribution is given by 
\begin{equation} \label{eq:PSD_orig}
 	K(\mathbf{w}; \bm{\mu}, \kappa) = (4\pi)^{-\eta} \frac{\Gamma(2\eta+\kappa)}{\Gamma(\eta+\kappa)} \left[\tfrac{1}{2}(1 + \mathbf{w}^\top\bm{\mu})\right]^\kappa,
\end{equation}
where $\bm{\mu}\in\dsphere$ is the mean direction, $\kappa\ge0$ is the bandwidth parameter, $\eta=(d-1)/2$, and $\Gamma(\cdot)$ is the gamma function. The dot product of two unit vectors is the cosine of the angle between them. Therefore, the term $\mathbf{w}^\top\bm{\mu} \in [-1,1]$, defines the cosine of the arc length between $\mathbf{w}$ and $\bm{\mu}$, and plays an analogous role to the Euclidean distance $\|\mathbf{x} -\bm{\mu}\|_2 \in [0, \infty)$ for isotropic kernels on $\RR^d$. 

For KD models, the bandwidth parameter $\kappa$ determines the balance between over-smoothing and over-fitting; it's optimisation is discussed below. The PS distribution with $\kappa=0$ is the uniform distribution on $\dsphere$. As $\kappa$ increases, the distribution becomes more concentrated about the mean direction. For larger sample sizes, using higher bandwidth parameters tends to result in a better resolution of complex distributional features. However, this can lead to numerical overflow problems when evaluating the gamma functions in \eqref{eq:PSD_orig}, since $\Gamma(z) \sim  \sqrt{2\pi z} (z/e)^{z}$ as $z\to\infty$. A more numerically stable expression is 
\begin{align*} \label{eq:PSD}
 	K(\mathbf{w}; \mu, \kappa) = (4\pi)^{-\eta} \exp \left[ \kappa \log(z) + \log\left(\Gamma(2\eta+\kappa)\right) - \log\left(\Gamma(\eta+\kappa)\right)\right],
\end{align*}
where $z=\tfrac{1}{2} (1 + \mathbf{w}^\top\bm{\mu}) \in [0,1]$ and hence $\log(z)\in(-\infty,0]$. We also have $\log(\Gamma(2\eta+\kappa)) - \log(\Gamma(\eta+\kappa)) \sim \eta \log(\kappa)$ as $\kappa\to\infty$. The log gamma function $\log(\Gamma(\cdot))$ can be computed directly in most software without having to evaluate the gamma function, thus avoiding overflow issues. 

\subsubsection{Bandwidth optimisation}
There are various ways in which the bandwidth parameter for the KD model can be optimised, as discussed in \cite{ley2017modern}. Here we use a cross-validation scheme to minimise the negative log-likelihood (NLL) for a hold-out sample. This is the most computationally expensive part of the method -- for each test point, the kernel must be evaluated for each observation in the training sample. For a leave-one-out cross-validation scheme, this entails $(n-1)^2$ evaluations of the kernel:
\begin{equation} \label{eq:KD_NLL}
    \mbox{NLL}(\kappa) = - \sum_{i=1}^n \log\left(\hat{f}_{\ne i}(\mathbf{w}_i)\right),
\end{equation}
where $\hat{f}_{\ne i}(\mathbf{w}_i)$ is the KD estimate \eqref{eq:KD} of the density from all observations except $\mathbf{w}_i$, evaluated at $\mathbf{w}_i$ (i.e., the predictive likelihood). However, as the NLL is effectively an expectation over $\mathbf{W}$, we can take a stochastic approach and estimate the NLL from a smaller sample of size $m$, chosen at random from the observations, so that we only need $(n-1)m$ kernel evaluations. This is similar to the approach used in stochastic gradient descent (see below) in machine learning, where a small batch of observations is used to approximate the true gradient of the cost function. 

When observations are serially correlated, the estimate of $\hat{f}_{\ne i}(\mathbf{w}_i)$ will be derived, in part, from observations that are correlated with $\mathbf{w}_i$, namely $\{...,\mathbf{w}_{i-2}, \mathbf{w}_{i-1}, \mathbf{w}_{i+1}, \mathbf{w}_{i+2},...\}$. To obtain an unbiased estimate of the predictive likelihood using a cross-validation scheme, for each point $\mathbf{w}_i$ at which we want to calculate the predictive likelihood, we need to leave out observations around $\mathbf{w}_i$ which are correlated with it, say $\mathbf{w}_{i-k},...,\mathbf{w}_{i+k}$ for some fixed $k$. 
 
\begin{figure}[ht]
\centering
\includegraphics[scale=0.7]{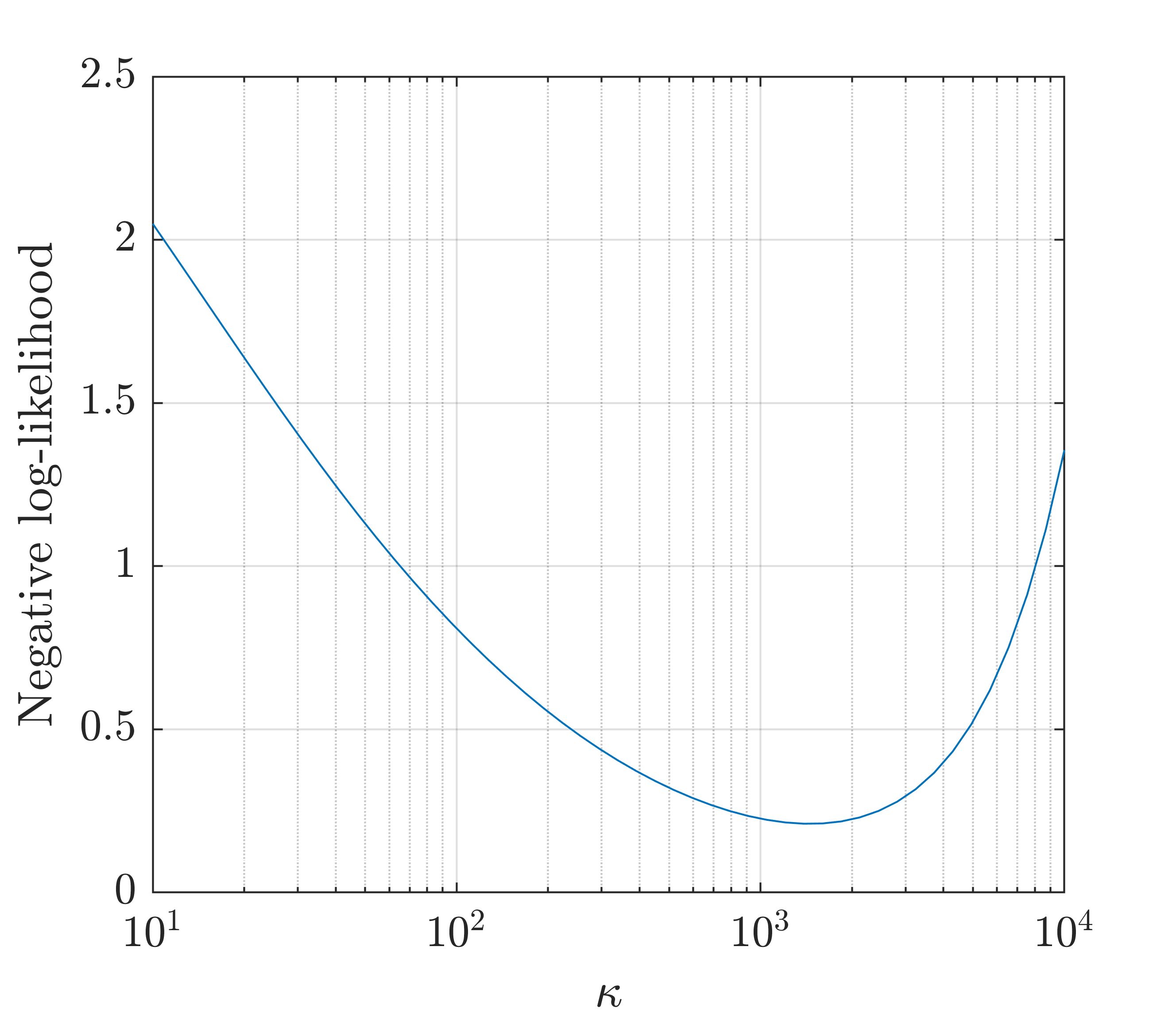}
\caption{Optimisation of bandwidth parameter $\kappa$ for kernel density estimate of the angular distribution, in terms of the minimum predictive negative log-likelihood.}
\label{fig:KD_bandwidth}
\end{figure}

In the example presented in Section \ref{sec:application}, we found that using $m=1000$ prediction points for estimating the NLL gave smooth and repeatable results for our sample size of $n=271,704$ hourly observations, requiring $\approx 271\times 10^6$ evaluations of the kernel (a speed up of $\approx 271$ times compared to the full cross-validation scheme). We also set $k=48$, corresponding to a two-day exclusion around each point used to calculate the predictive NLL. Computations of the NLL for 50 logarithmically-spaced test values of $\kappa\in[10^1,\, 10^4]$ took approximately 5 minutes on a laptop with an Intel Core i7-1355U 1.7GHz processor. The results are shown in \autoref{fig:KD_bandwidth}, which indicates an optimal value of $\kappa\approx 1200$. 

\subsubsection{Simulation from a kernel density estimate}
Once the value of the bandwidth parameter has been optimised, it is straightforward to simulate from the KD model. As mentioned above, the KD model is effectively a mixture model, with one component corresponding to each observation, centred at the location of the observation. Therefore, to generate a random sample from the KD model requires two steps. First, we sample an observed angle $\mathbf{w}_i$ at random (i.e., a random integer $i\in\{1,...,n\}$). This corresponds to choosing one component from the mixture model at random. Secondly, we generate a random value from the kernel with mean direction $\mathbf{w}_i$ and bandwidth $\kappa$. As described in Appendix \ref{app:PSD_simulation}, simulation from the PS kernel on $\dsphere$ requires generating a beta random variable and $d-1$ standard normal random variables, and applying a rotation. Computationally this is very efficient. For the example presented in Section \ref{sec:application}, generating a sample of size $2.7\times10^7$ (100 times the observed sample size) took approximately 10~s on a laptop with specifications mentioned above.
    
\subsection{Conditional radial modelling} \label{sec:rad_dens}
Viewing the angular variable $\mathbf{W}$ as a `covariate' for the radial variable $R$, inference for the SPAR model is analogous to a non-stationary univariate POT analysis, for which many parametric and semi-parametric approaches have been proposed \cite{chavez2005, Randell2016, Youngman2019, Zanini2020, Barlow2023}. Non-stationary POT modelling can be performed using generalised Pareto (GP) regression (i.e., modelling GP parameters as functions of covariates), which proceeds by first estimating the threshold function, $u(\mathbf{w})$, via a quantile regression procedure, and then estimating the GP scale and shape parameter functions, $\sigma(\mathbf{w})$ and $\xi(\mathbf{w})$ respectively, via likelihood-based inference procedures. The choices of the functional forms for $u(\mathbf{w})$, $\sigma(\mathbf{w})$, and $\xi(\mathbf{w})$ determine the flexibility of the overall model. As the dimension $d$ grows, these mappings become increasingly complex, and models that represent the functions via semi-parametric models (such as the splines used in initial work with the SPAR model \cite{murphy2024inference} and other similar angular-radial approaches \cite{simpson2024estimating, majumder2023semiparametric}) become increasingly computationally-demanding to estimate. Consequently, we adopt a deep learning approach, whereby the threshold and GP parameter functions are represented using artificial neural networks (ANN). For details on GP regression with deep learning methods, see \cite{Richards2024}. 

Although the representation of the GP parameter functions via ANNs differs from previous approaches in the SPAR framework, the `loss function' used for optimisation of the model is the same. That is, we estimate the threshold and GP parameter functions that minimise the relevant loss functions evaluated on a hold-out dataset (in the case of the GP parameters, the loss function is the negative log-likelihood), as described in Section \ref{sec:nn_fitting}. This work builds upon the approach of \cite{murphy2024deep}, who use deep learning to estimate the extremal dependence structure of random vectors on standardised marginal scales via a similar angular-radial decomposition. In contrast, our approach does not require marginal transformation, and is thus not subject to marginal estimation uncertainty.

\subsubsection{Neural network representation of conditional radial parameters} \label{sec:deep_learn}
Several recent approaches have used deep learning for GP regression; see, e.g.,  \cite{wilson2022deepgpd, richards2023insights, cisneros2024deep, pasche2024neural}. In these approaches, ANNs are used to model the relationships between covariates and GP parameter and threshold functions. Our approach is analogous, with the covariates taken to be angles on the hypersphere. Note that the hypersphere is compact (i.e., closed and bounded), and that this is a desirable property for extrapolation in deep learning. 

Here, we design two neural network models; one for the threshold function $u(\mathbf{w})$ and one for the parameter vector $(\nu(\mathbf{w}),\xi(\mathbf{w}))$, where $\nu(\mathbf{w}) = \sigma(\mathbf{w})( \xi(\mathbf{w})+1)$ is the modified scale parameter. Unlike $\sigma(\mathbf{w})$, the modified scale parameter $\nu(\mathbf{w})$ is orthogonal to $\xi(\mathbf{w} )$, which helps to mitigate numerical instabilities during model fitting \cite{cox1987parameter, chavez2005, tendijck2024practical};  \cite{pasche2024neural} show that this is particularly helpful when fitting deep GP regression models. Both $u(\mathbf{w})$ and $(\nu(\mathbf{w}),\xi(\mathbf{w}))$ are modelled by multi-layer perceptrons (MLPs), which are a standard class of fully-connected ANN that compose multiple layers of `neurons' \cite{goodfellow2016deep}. Each neuron passes a linear combination of input variables through a nonlinear `activation function', and the output is then passed to the subsequent layer; detailed discussions and illustrative figures can be found in \cite{Richards2024}, and an example schematic for an MLP representation of $(\nu(\mathbf{w}),\xi(\mathbf{w}))$ is presented in \autoref{fig:schematic}. Inference for the parameter functions involves estimating the linear coefficients (the `weights' and `biases') in each neuron of the corresponding MLP. 

Prior to inference, the architecture of the MLP must be defined, i.e., the number of hidden layers, denoted by $L$, the number of neurons in each hidden layer, denoted by $h_1, \ldots, h_L$, and the type(s) of activation function(s). The resulting set of estimable parameters for the MLP contains all of the weights and biases in each hidden layer, as well as the final $(L+1)$-th layer; we denote this by $\mathcal{W}:=\left\{\left(a^l, b^l\right) ; l=1, \ldots, L+1\right\}$, with weights $a^l \in \mathbb{R}^{h_l \times h_{l-1}}$ and biases $b^l \in \mathbb{R}^{h_l}$. Note that the estimable sets of parameters differ between the MLPs for $u(\mathbf{w})$ and  $(\nu(\mathbf{w}),\xi(\mathbf{w}))$; we denote their respective parameter sets by $\mathcal{W}_u$ and $\mathcal{W}_{(\nu,\xi)}$. For both MLPs, we take all hidden layer activation functions to be the rectified linear unit function, $\text{ReLU}(\mathbf{x})=(\max\{x_1,0\},\max\{x_2,0\},\dots)$. The final layers of the MLPs make use of an exponential transformation to ensure that the scale $\nu(\mathbf{w})$ and threshold $u(\mathbf{w})$ are strictly positive, i.e., $\nu(\mathbf{w})>0,u(\mathbf{w})>0$ for all $\mathbf{w}$. For numerical stability, we also ensure that $\xi(\mathbf{w})$ satisfies $\xi(\mathbf{w})\in(-0.5,0.1)$ for all angles $\mathbf{w}$. Selection of the remaining tuning parameters is discussed in Section~\ref{sec:arch}. 

\begin{figure}
\centering
\begin{tikzpicture}[x=2.4cm,y=1.2cm]
  \readlist\Nnod{3,4,4,2} 
  \readlist\Nstr{d,h_{\prev},} 
  \readlist\Cstr{w,w^{(\prev)},\nu(\mathbf{w})} 
  \def\yshift{0.55} 
  
  \foreachitem \N \in \Nnod{
    \def\lay{\Ncnt} 
    \pgfmathsetmacro\prev{int(\Ncnt-1)} 
    \foreach \i [evaluate={\c=int(\i==\N); \y=\N/2-\i-\c*\yshift;
                 \x=\lay; \n=\nstyle;
                 \index=(\i<\N?int(\i):"\Nstr[\n]");}] in {1,...,\N}{ 
       \ifnum \lay<\Nnodlen
         \node[node \n] (N\lay-\i) at (\x,\y) {$\strut\Cstr[\n]_{\index}$};
      \fi
       \ifnum \lay=\Nnodlen
       \ifnum \i=1
         \node[node \n] (N\lay-\i) at (\x,\y) {$\nu(\mathbf{w})$};
         \fi
           \ifnum \i=2
         \node[node \n] (N\lay-\i) at (\x,\y) {$\xi(\mathbf{w})$};
         \fi
      \fi
      
      \ifnumcomp{\lay}{>}{1}{ 
        \foreach \j in {1,...,\Nnod[\prev]}{ 
          \draw[white,line width=1.2,shorten >=1] (N\prev-\j) -- (N\lay-\i);
          \draw[connect] (N\prev-\j) -- (N\lay-\i);
        }
        \ifnum \lay=\Nnodlen
          \draw[connect] (N\lay-\i) --++ (0.5,0); 
        \fi
      }{
        \draw[connect] (0.5,\y) -- (N\lay-\i); 
      }
      
    }
      \ifnum \lay<\Nnodlen
        \path (N\lay-\N) --++ (0,1+\yshift) node[midway,scale=1.6] {$\vdots$}; 
        \fi
}
 
  \node[above=3,align=center,mydarkgreen] at (N1-1.90) {Input\\[-0.2em]layer};
  \node[above=2,align=center,mydarkblue] at (N2-1.90) {Hidden\\[-0.2em]layer 1};
  \node[above=2,align=center,mydarkblue] at (N3-1.90) {Hidden\\[-0.2em]layer 2};
  \node[above=3,align=center,mydarkred] at (N4-1.90) {Output\\[-0.2em]layer};
 
\end{tikzpicture}
\caption{Example schematic of a multi-layered perceptron (MLP) model for the GP parameters, with $L=2$ hidden layers. The inputs are the components of the angle $\mathbf{w} = (w_1,w_2,...,w_d)$ and the outputs are the GP parameter functions, $(\nu(\mathbf{w}), \xi(\mathbf{w}))$.}
\label{fig:schematic}
\end{figure}

\subsubsection{Estimating the neural network parameters} \label{sec:nn_fitting}

To obtain estimates of the MLP parameter sets, $\mathcal{W}_u$ and $\mathcal{W}_{(\nu,\xi)}$, we optimise specified loss functions. Suppose that we have a set of radial and angular observations $\{(r_i,\mathbf{w}_i); i=1,...,n\}$. Recall from Section \ref{sec:SPAR_def} that the threshold $u(\mathbf{w})$ is taken to be the quantile of $R|(\mathbf{W}=\mathbf{w})$ at exceedance probability $\zeta$. We thus can estimate $u(\mathbf{w})$ (and its corresponding parameter set $\mathcal{W}_u$) using quantile regression techniques \cite{koenker2017handbook}. In this case, an appropriate loss function for $u(\mathbf{w})$ is the tilted loss,
\begin{equation}
\label{eqn:q_ll}
    \mathcal{L}_u(\mathcal{W}_u) := \sum_{i=1}^n \rho_{1-\zeta}\left\{r_i-u\left(\mathbf{w}_i\right)\right\},
\end{equation}
where $\rho_{\alpha}(t):=t(\alpha-\mathbbm{1}\{t<0\})$ for indicator function $\mathbbm{1}$ and where dependency of $u(\mathbf{w})$ on $\mathcal{W}_u$ has been suppressed from notation. 

After estimation of $u(\mathbf{w})$, we define
$I_u := \{i\in\{1,...,n\} : r_i > u (\mathbf{w}_i) \}$ as the set of indices of radial threshold exceedances. The MLP that defines the GP parameter functions can be considered as a `conditional density estimation network', with the negative log-likelihood function used for optimisation; see, e.g., \cite{neuneier1994estimation, rothfuss2019conditional}. In this case, we perform maximum likelihood estimation, with loss 
\begin{equation} \label{eqn:gp_ll}
   \mathcal{L}_{GP}(\mathcal{W}_{(\nu,\xi)}):= - \sum_{i \in I_u} \log \left[ f_{\rm GP}\left(r_i - u(\mathbf{w}_i); \xi(\mathbf{w}_i) , \frac{\nu(\mathbf{w}_i)}{\xi(\mathbf{w}_i) + 1} \right) \right].
\end{equation}

We first train the ANN for the threshold, and use this to define the set of threshold exceedances. The ANN for the GP parameters is trained afterwards. Optimisation of both losses, \eqref{eqn:q_ll} and \eqref{eqn:gp_ll},  proceeds via stochastic gradient descent and the ADAM algorithm \cite{kingma2014adam}. To mitigate overfitting, data are split into training (80\%) and validation (20\%) sets, with the latter used to check for parameter convergence. We refer the reader to \cite{murphy2024deep} for a more detailed overview of the fitting procedure.  

As noted in Section~\ref{sec:theory}, when the GP shape parameter $\xi(\mathbf{w})$ is negative, the distribution of $R|(\mathbf{W}=\mathbf{w})$ has a finite upper endpoint. Training of a deep GP regression model which permits negative shape parameter values can be computationally troublesome; see discussion by \cite{richards2022regression}. At a given angle $\mathbf{w}_i$, if $\xi(\mathbf{w}_i)<0$ and the radial observation $r_i$ exceeds the upper endpoint, i.e., $r_i > u(\mathbf{w}_i) - \sigma(\mathbf{w}_i)/\xi(\mathbf{w}_i)$, then the loss function in \eqref{eqn:gp_ll} will evaluate to a infinite value. Consequently, the loss surface over which we optimise $\mathcal{W}_{(\nu,\xi)}$ is highly irregular, and iterative gradient descent methods (like ADAM) may have trouble finding global maxima, or may predict out-of-sample parameter estimates that are infeasible, i.e., the loss is infinite. To circumvent these issues during training, we initialise the MLP to ensure that the shape parameter function $\xi(\mathbf{w})$ is non-negative for all angles $\mathbf{w}$; in this way, at the outset of the training procedure, the loss function is guaranteed to be finite for all $w_i,i=1,\dots,n$. Then, if the gradient descent optimisation algorithm produces infinite loss values during training, we restart training (from the last iteration with finite loss values) with a smaller learning rate. In our application we found that the fully-trained MLP provided negative values of $\xi(\mathbf{w})$ for all $\mathbf{w}\in\dsphere$. We note that the issue of infinite loss occurs in other likelihood-based inference procedures for GP regression, and is not a problem specific to MLPs.

\subsubsection{Selection of hyperparameters} \label{sec:arch}

The two key choices for the proposed inference scheme are the architecture of the neural network and the threshold non-exceedance probability, $\zeta$. There is no `best practice' for this selection within the deep GP regression literature \cite{Richards2024}, and the appropriate architecture is likely to be domain specific; see \cite{goodfellow2016deep}. The neural network architecture determines the flexibility of the model. Selecting a model with more hidden layers and more neurons results in higher flexibility, but at the cost of increased parameter variability and computational expense. Moreover, the optimal architecture is intrinsically linked to the optimal $\zeta$, as having more exceedances, i.e., more data, allows for estimation of a more flexible model. In the spirit of parsimony, we wish to select the simplest model possible while still capturing the observed variability in the threshold and GP parameter functions over the angular domain. 

To select our `optimal' architecture and $\zeta$ for the application detailed in Section~\ref{sec:application}, we perform a grid-search over architecture choices and a sequence of threshold exceedance probabilities in the range $0.0125\le \zeta\le 0.25$ at intervals of $0.0125$. For each configuration, we examine a range of goodness-of-fit diagnostic plots, discussed in Section~\ref{sec:radial_assess}. The optimal architecture and $\zeta$ is then chosen as those values which visually provide the best model diagnostics. For both the threshold $u(\mathbf{w})$ and GP excess models, a simple neural network architecture is preferable: $L=3$ hidden layers, with $h_l=16$ neurons per layer. This results in two MLPs, each comprising approximately 650 estimable parameters; inference for these models is not computationally demanding, and can be conducted on a standard laptop

\begin{figure}
\centering
\includegraphics[scale=0.75]{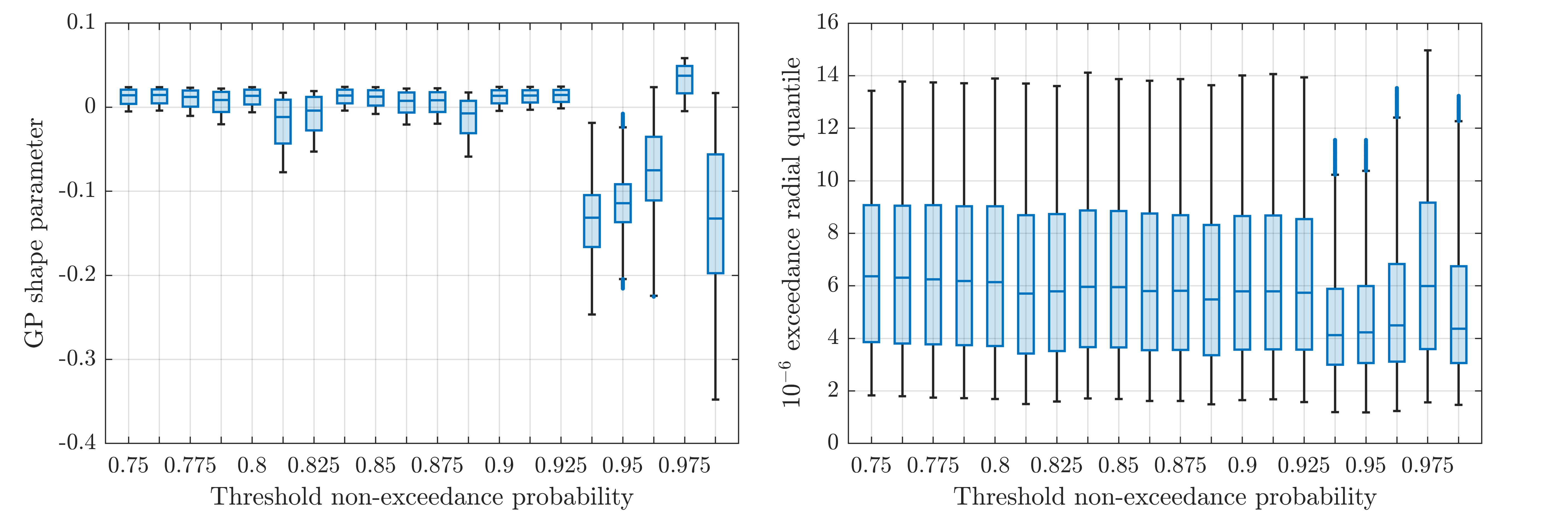}
\caption{Box-plots of GP shape parameter estimates (left) and conditional radial quantile at exceedance level $10^{-6}$ (right) at observed angles for various threshold non-exceedance probabilities, $1-\zeta$.}
\label{fig:threshold}
\end{figure}

As with univariate POT models, selecting a suitable threshold for our model is critical. Too low a threshold will result in the asymptotic arguments motivating the use of the GP model not being applicable, causing bias; whereas too high a threshold will result in too few observations for fitting, resulting in higher variance. Thus, alongside the aforementioned visual diagnostics, we further justify the choice of $\zeta$ using a process of threshold selection which is directly analogous to that in univariate problems. For a fixed choice of architecture (as given above), we fit the model for a sequence of $\zeta$ values and check for stability of estimates of the GP shape parameter and conditional radial quantiles. The optimal exceedance probability is then selected as the smallest value of $\zeta$ for which inferences are approximately stable for $\zeta_0>\zeta$ (allowing for sampling uncertainty). As the GP shape parameter and radial quantiles are non-stationary, we need to assess how their values vary with threshold for a range of different angles. To give a diagnostic that is weighted towards regions with higher angular density, we consider the distribution of shape parameter and quantile estimates over all observed angles. The information is summarised as box plots, as shown in \autoref{fig:threshold}. In our application, we found that $\zeta=0.1$ gave the best results. 

We remark that while our selected architecture works well for our application, we do not advocate the general use of these hyperparameters. Instead, we recommend that practitioners who apply our framework perform a similar grid-search, and use post-fit diagnostics to select the optimal architecture.

\section{Application to five-dimensional problem} \label{sec:application}
\subsection{Dataset} \label{sec:data}
In this section, we consider the application of the SPAR model to a hindcast dataset consisting of 31-years of wind and wave variables from 01/01/1990 to 31/12/2020, for a site in the Celtic Sea, off the south-west coast of the UK, which has been identified for development of floating wind farm projects. The data was produced using a one-way coupled wave-current model, described in \cite{mackay2022joint}. At the site considered, the mean water depth is 53~m with a 7~m tidal range. The dataset consists of hourly values of significant wave height $(H_s)$, mean wave period $(T_m)$, mean wave direction $(\theta_{wave})$, hourly mean wind speed at 10 m above sea level $(U_{10})$, and wind direction $(\theta_{wind})$. Both wind and wave directions are defined as the directions in which the winds or waves are `going to', and directions are defined clockwise from North. These variables all influence the motion and loading of floating wind turbines, and understanding their joint extremes is important for design. Since directional variables are periodic, it does not make sense to talk about `extreme directions'. Instead, we work with the x- and y-components of wave height and wind speed, defined as $H_x = H_s \cos(\theta_{wave})$, $H_y = H_s \sin(\theta_{wave})$, $U_x = U_{10} \cos(\theta_{wind})$, and $U_y = U_{10} \sin(\theta_{wind})$. 

\begin{figure*}[p!]
\centering
\includegraphics[scale=0.75]{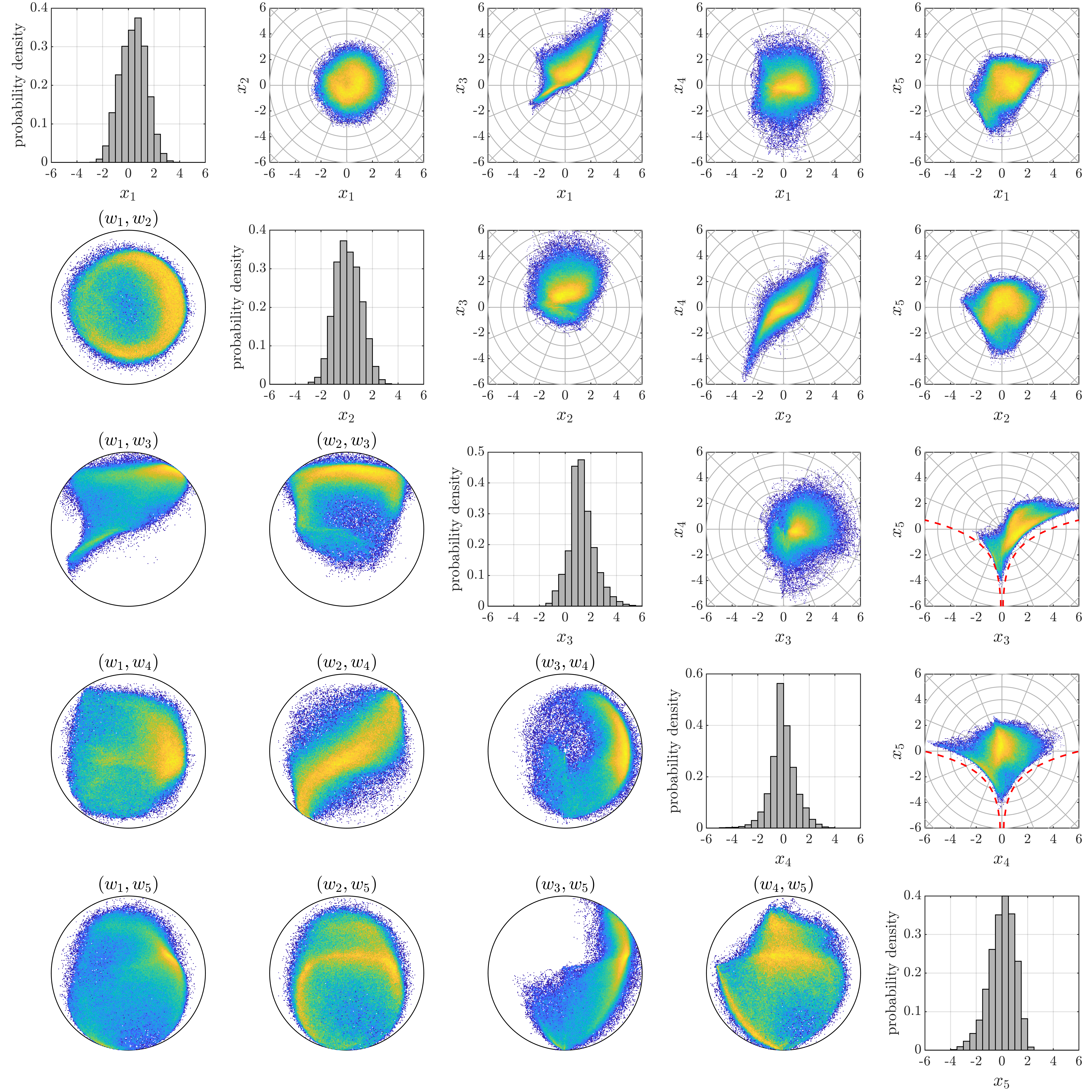}
\caption{Upper right plots: Empirical joint densities of pairs of normalised observations. Dashed red lines in plots of $(x_3,x_5)$ and $(x_4,x_5)$ are lines of constant wave steepness $s=0.08$. Plots on diagonal are histograms of each normalised variable. Lower left plots: Empirical density of pairs of angular components of observed data (yellow = high density, blue = low density). The normalised variables $(X_1 , ..., X_5)$ correspond to normalised $(U_x,U_y,H_x,H_y,\log(T_m))$. }
\label{fig:scatters}
\end{figure*}

Unlike many classical modelling approaches, the SPAR approach extrapolates in all directions, allowing one to perform inference in any extreme region of interest. When variables have a defined lower bound at zero, as is the case for many physical quantities, the SPAR model should be able to infer this directly from the data, and these physical limits should correspond to the upper bounds of the GP model for the radial variable at the relevant angles. However, inferences at endpoints are highly uncertain since they correspond to zero exceedance probability, and consequently the model may infer a slightly negative bound in some directions. It is therefore safer to work with variables that do not have hard lower bounds at zero, so that uncertainties in the radial endpoint do not result in estimates that are not physically possible. The directional variables $H_x$, $H_y$, $U_x$, and $U_y$ are all defined on $(-\infty,\infty)$, although clearly they will have some upper and lower bounds due to physical constraints (in general, we would expect the region of variable space in which the density is non-zero to be bounded, due to physical limits). For the period variable, we address the problem of the lower bound by defining $L_T=\log(T_m) \in(-\infty,\infty)$ and using this variable in the model. 

\subsection{Normalisation and choice of origin}
As different physical variables have different scales, we normalise each variable by its standard deviation. Without this, angles would tend to be clustered in the plane of whichever variable has the largest scale (wind speed in this case). We also need to define an origin in order to transform to polar coordinates. In previous work using SPAR \cite{mackay2025spar, murphy2024inference}, the origin was defined at the mean of each variable. In our application, more care must be taken when defining an appropriate origin. For the model to provide a useful description of the extremes at all angles, the support of the density, $\mathrm{supp}(f_{\mathbf{X}}) := \{\mathbf{x}\in\mathbb{R}^d : f_{\mathbf{X}}(\mathbf{x}) > 0\}$, must be star-shaped with respect to the chosen origin \cite{hansen2020}. That is, given an origin $\mathbf{x}_0 \in \mathrm{supp}(f_{\mathbf{X}})$ and any point $\mathbf{x} \in \mathrm{supp}(f_{\mathbf{X}})$, the line segment from $\mathbf{x}_0$ to $\mathbf{x}$ is contained in $\mathrm{supp}(f_{\mathbf{X}})$. Under this assumption, all rays from the origin reach the `edges' of the distribution without passing through regions which have zero density. In this way, the data-cloud has a well-defined `inside' and `outside', with the `outer' region considered `extreme', and the GP-based representation of the radial component in this region is reasonable. This assumption can be verified by checking plots of the density of observations along various rays from the origin, e.g., histograms of the observed radial variable within small angular ranges. However, as discussed further below, in higher dimensional spaces, a large number of angles are required to obtain a reasonable coverage of the surface of $\mathbb{S}^{d-1}$, and visual inspection of plots of the radial density over each angular range is time consuming. Ultimately a data-driven approach for selecting an optimal choice of origin would be preferable.

In this example, we use knowledge of the physical processes to define an appropriate origin. Firstly, wave breaking limits the maximum possible wave height for a given wave period, with the limit related to wave steepness given by $s=2\pi H_s/(g T_m^2)$. In the three-dimensional subspace containing the variables $(H_x,H_y,T_m)$, this results in a conical-shaped bound on the data, centred along the axis $H_x=H_y=0$, with the radius of the cone given by $H_{s,max} = s_{max} g T_m^2 /(2\pi)$, where $s_{max}$ is the limiting steepness. The value of $s_{max}$ depends on the water depth and wind speed (among other factors), but the maximum value was found to be around $s_{max}\approx 0.08$ for our dataset (consistent with values in offshore standards, e.g. \cite{DNVC205}) . This conical shape to the distribution (see Figures~\ref{fig:scatters} and \ref{fig:height_period_direction}) suggests that an appropriate choice of origin should be somewhere on the axis $H_x=H_y=0$. Due to the physical dependence of wave height on wind speed, we also locate the origin at $U_x=U_y=0$. The choice of origin for $L_T$ is somewhat arbitrary, but experimentation showed that using the mean value of $L_T$ gave satisfactory results. 

The angular and radial variables are therefore defined with respect to the normalised variables given by {$X_1 = U_x / \mathrm{STD}(U_x)$}, {$X_2 = U_y / \mathrm{STD}(U_y)$}, $X_3 = H_x / \mathrm{STD}(H_x)$, $X_4 = H_y / \mathrm{STD}(H_y)$, and $X_5 = (L_T - \mathrm{mean}(L_T)) / \mathrm{STD}(L_T)$, where $\mathrm{STD}(\cdot)$ denotes the standard deviation function. The pairwise relations between these normalised variables are shown in \autoref{fig:scatters}. A radial grid has been overlaid to illustrate that these two-dimensional projections are approximately star-shaped with respect to this origin. Lines of bounding steepness $s=0.08$ are shown in the plots of $(X_3,X_5)$ and $(X_4,X_5)$ as dashed lines. It can be seen that there is far less scatter in the variables close to these bounds due to the physical limitations. Another feature of the data that is evident is the strong positive correlation between the x- and y-components of the wave height and wind speed (i.e., the pairs $(X_1,X_3)$ and $(X_2,X_4)$). 

\subsection{Exploratory data analysis}
Before assessing the model fit, it is useful to consider various visualisations of the data, in order to understand how it is distributed over the five-dimensional space. The plots below the diagonal in \autoref{fig:scatters} show the empirical densities of pairs of angular components $(W_i,W_j)$, $i,j\in\{1,...,d\}$. By definition, these variables must fall within the unit circle. Any large gaps in the observed values indicate that it is not possible to fit the model at these angles. The objective of using the MLP model of the angular variation of the radial distribution is to estimate a relatively smooth variation with angle. The model should therefore be able to smooth over small angular ranges with no observations. However, the model is unlikely to be able to accurately estimate the behaviour of the radial component over large angular ranges where there is little or no data. 

\begin{figure}[ht]
\centering
\includegraphics[scale=0.7]{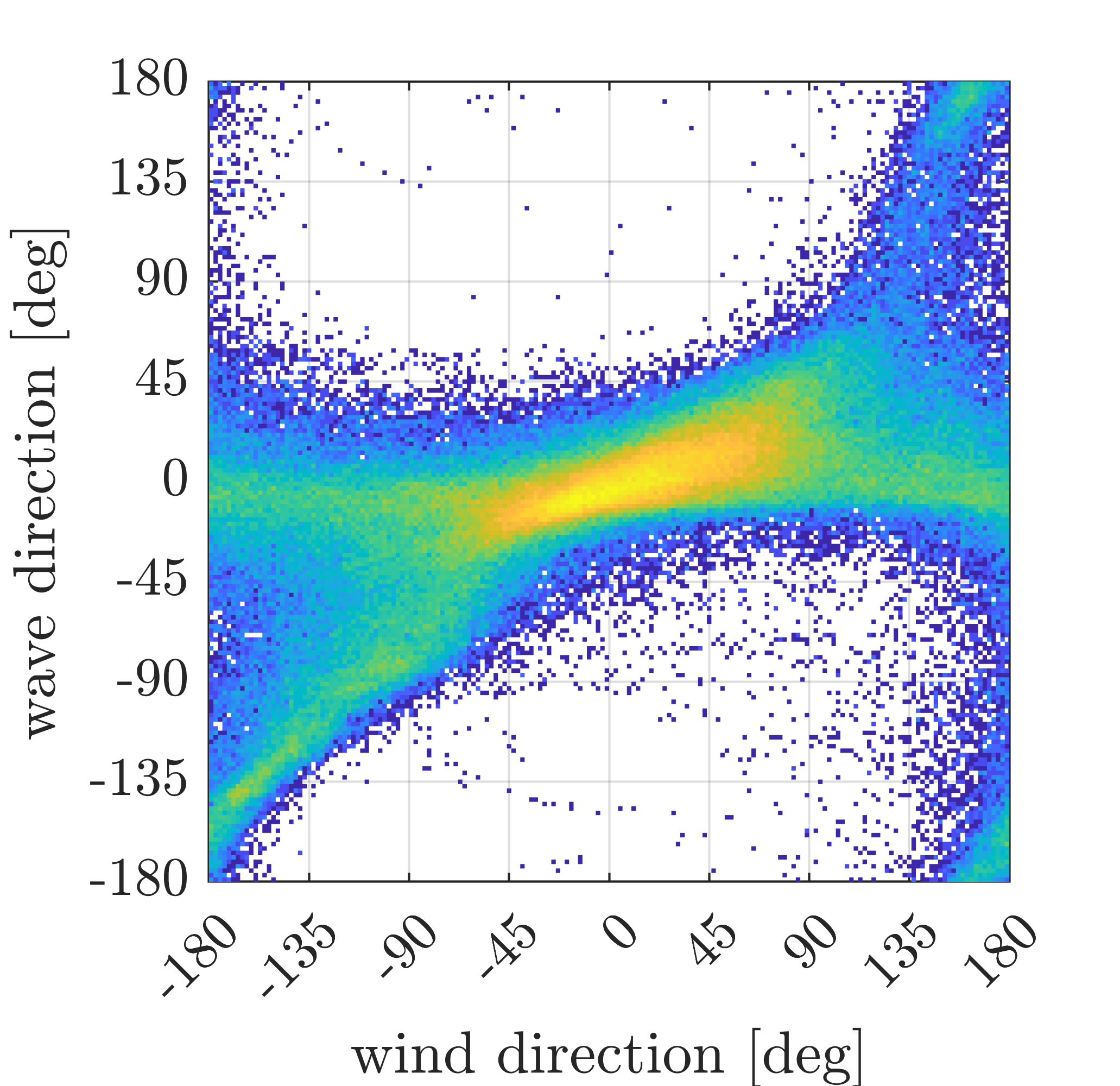}
\caption{Empirical joint density of wind direction and wave direction. (Blue = low density, yellow = high density).}
\label{fig:directions}
\end{figure}

Consider the joint occurrence of wave direction and wind direction, illustrated in \autoref{fig:directions}. Note that the wind direction is $\theta_{wind} = \mathrm{atan2}(W_1,W_2)$ and $\theta_{wave} = \mathrm{atan2}(W_3,W_4)$, where $\mathrm{atan2}(x,y)$ is the four-quadrant inverse tangent function. So the angles shown in \autoref{fig:directions} are a subset of $\mathbb{S}^3$ where $w_1^2+w_2^2 = w_3^2+w_4^2 = 1/2$ (known as the Clifford torus). As discussed above, wind and wave directions tend to be roughly aligned, although there is some scatter. However, there are large areas of the variable space where observations are sparse. So attempting to model the conditional joint distribution of three other variables (wind speed, wave height, wave period), let alone their joint extremes, is likely to be very challenging in these regions. 

To assess the variation in the density of angles around a circle, we could plot the number of observations within discrete angular ranges as a histogram. In higher dimensions, visualisation becomes more difficult, but a similar approach can be taken. We count the number of observations within a small angular range of a pseudo-regularly spaced grid of points on the sphere. It is not possible to define evenly-spaced points on the surface of a (hyper) sphere in three or more dimensions. To address this issue we define a set of points $U$ that are approximately regularly-spaced points on $\dsphere$, using the method described in Appendix \ref{app:L1L2_projection}. For each $\mathbf{u}\in U$ we count the number of observed angles with $\arccos(\mathbf{w}_i^\top \mathbf{u})<\theta_{max}$, where $\theta_{max}$ is some prescribed range. Although there may be some overlap between the ranges defined above, this analysis still gives an indication of how the density of angles varies over the sphere.  

\begin{figure}[ht]
\centering
\includegraphics[scale=0.7]{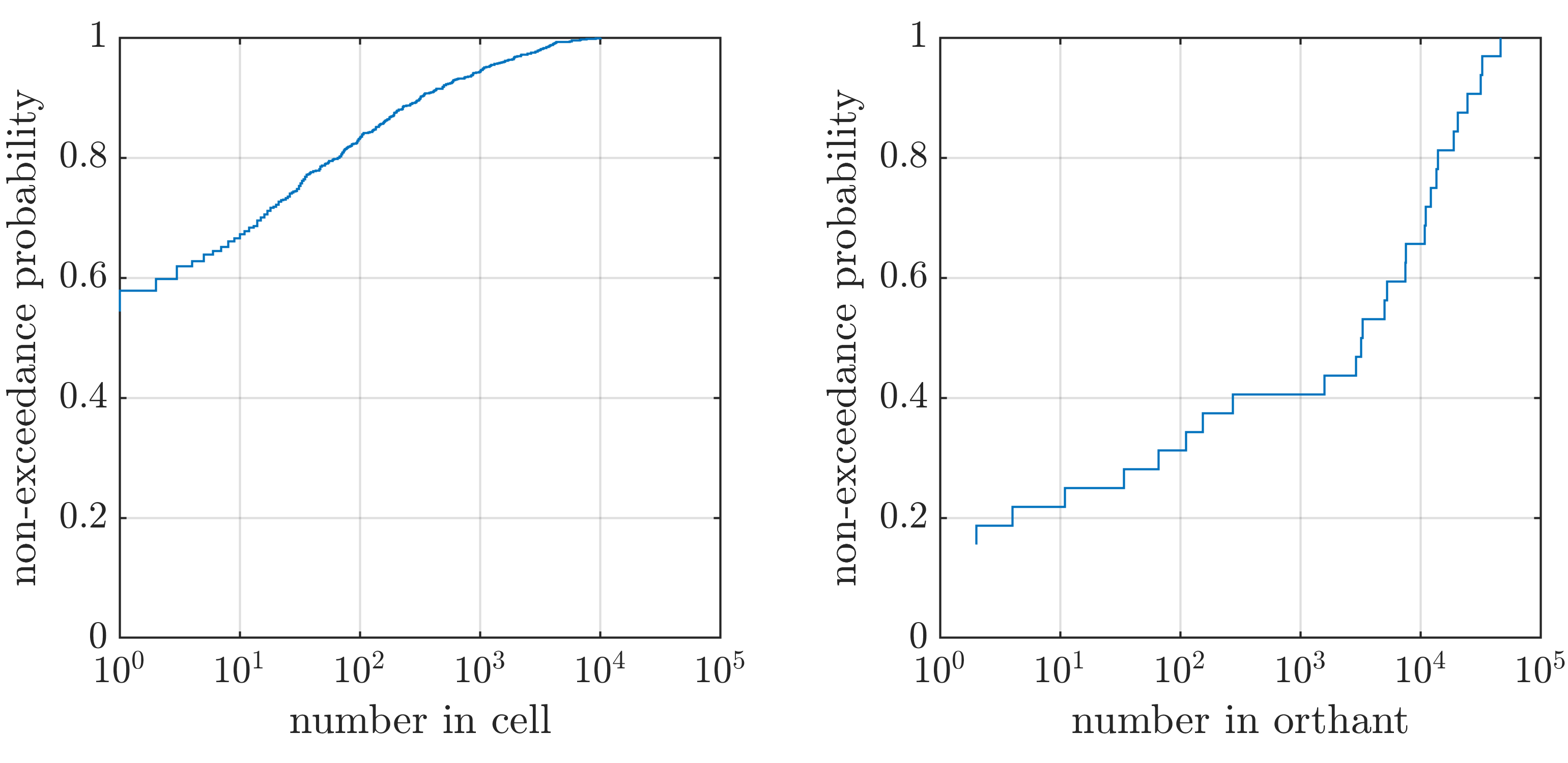}
\caption{Left: Empirical distribution of the number of points within a $15^\circ$ radius of each direction vector. Right: empirical distribution of the number of points in each of the 32 orthants of $\mathbb{R}^5$.}
\label{fig:ang_dens}
\vspace{10pt}
\includegraphics[scale=0.7]{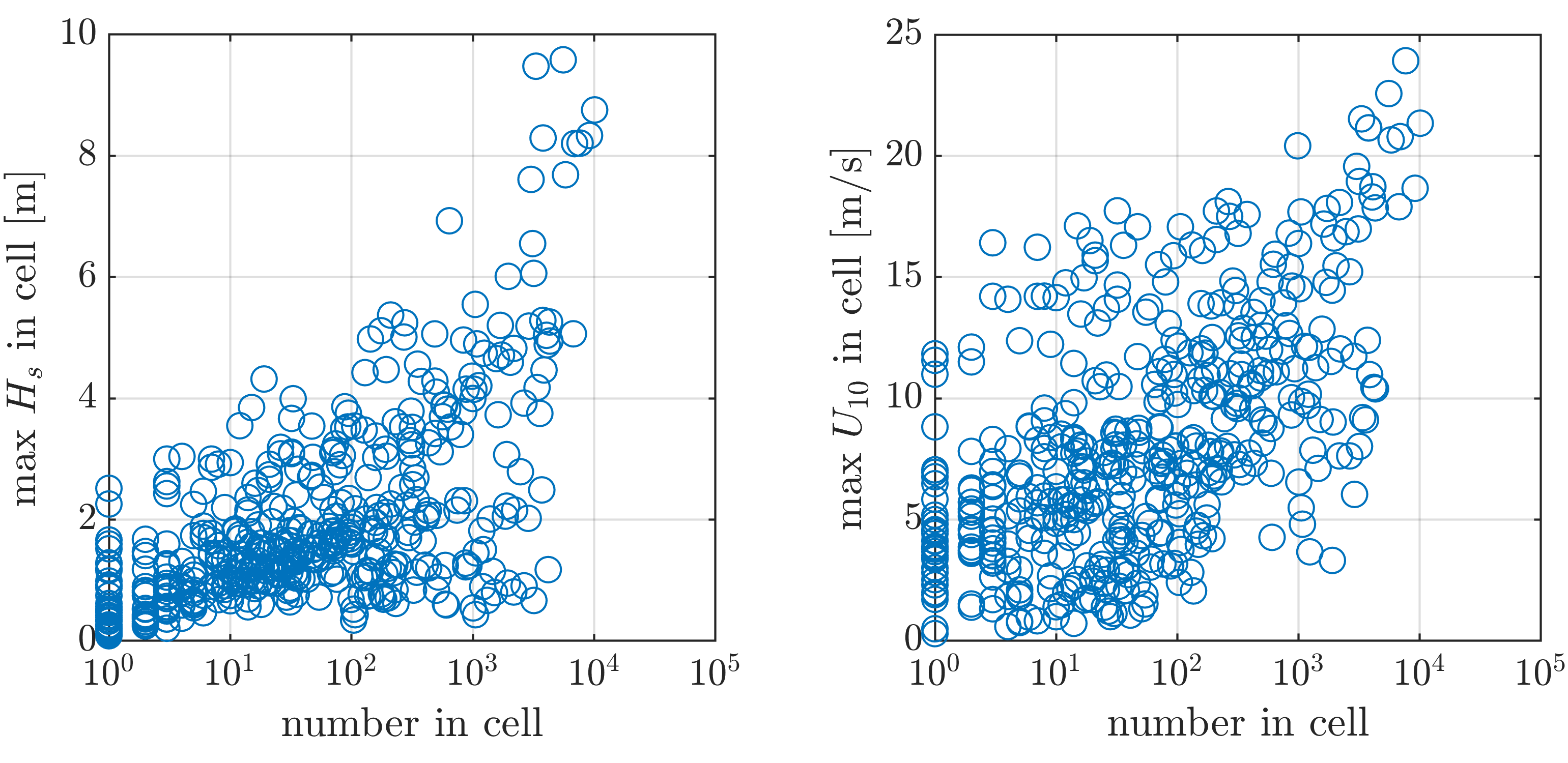}
\caption{Maximum observed $H_s$ and $U_{10}$ in each local angular cell  vs. number in cell.}
\label{fig:max_vs_num}
\end{figure}

\autoref{fig:ang_dens} shows the empirical cumulative distribution function (CDF) of the number of observations in a cell chosen at random, for a cell radius of $\theta_{max} = 15^\circ$ and a set of 1002 direction vectors (generated using $m=5$ points along each dimension -- see Appendix~\ref{app:L1L2_projection}). One feature that is evident is that over 50\% of angular cells contain no observations. This is due to the particular choice of origin, which was selected to meet the assumption of a star-shaped distribution. \autoref{fig:ang_dens} also shows the number of observations in each of the $2^5=32$ orthants in $\mathbb{R}^5$ (an orthant is the $d$-dimensional analogue of a quadrant of the plane). There are five orthants which contain no observations, and a further five which contain fewer than 100 observations (out of a total of 271,704 observations). The SPAR model estimates of the extremes in the orthants with little data will therefore be highly uncertain. However, \autoref{fig:max_vs_num} shows scatter plots of the maximum observed values of $H_s$ and $U_{10}$ in each of the 1002 angular cells against the corresponding number of observations in the cell, indicating that larger values of these variables tend to coincide with higher angular densities. The higher uncertainties associated with the lower occurrence regions should therefore have less effect on the global extremes. 

\begin{figure}[ht]
\centering
\includegraphics[scale=0.7]{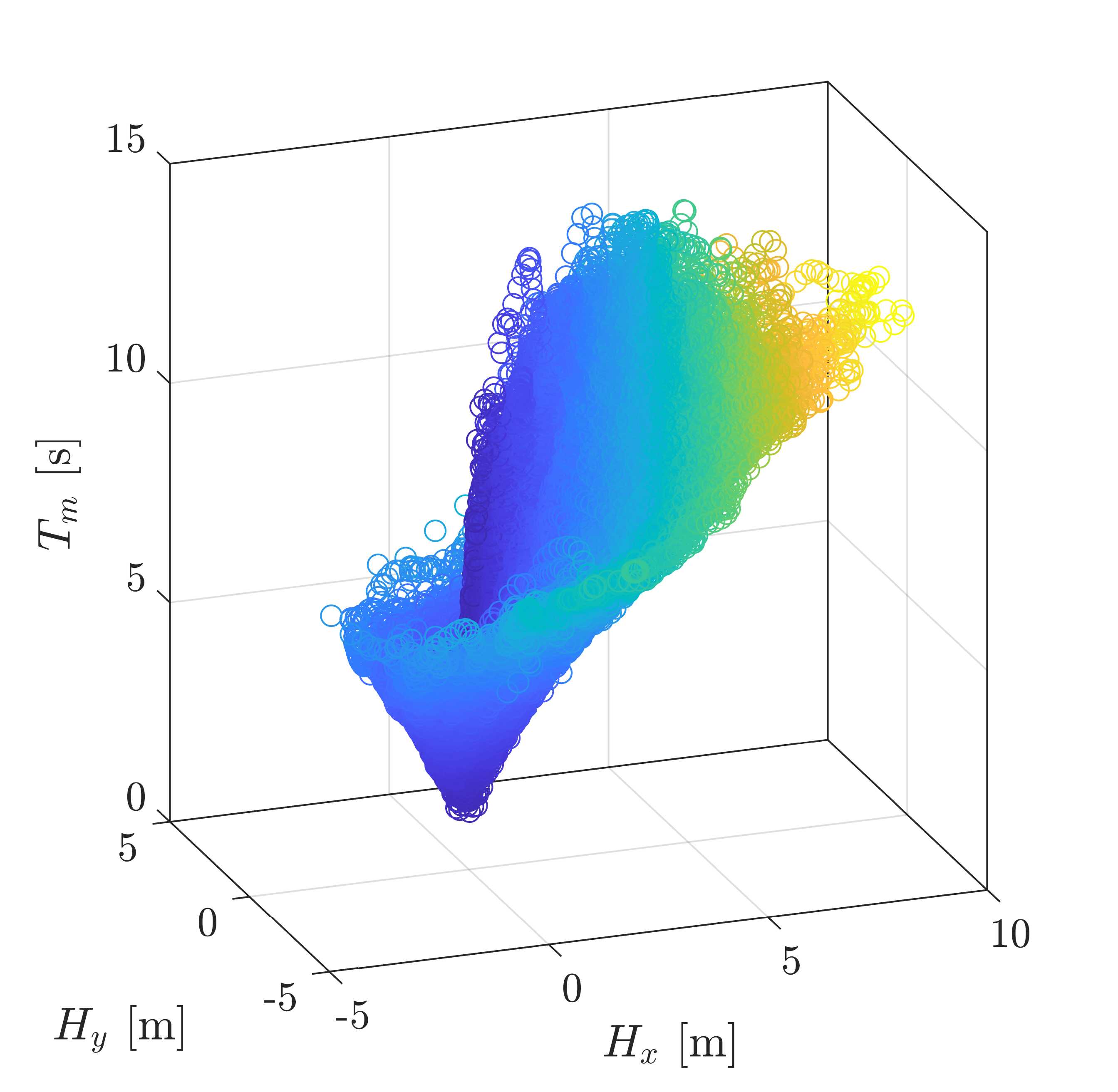}
\caption{Scatter plot of observations of x-y components of $H_s$ and $T_m$. Colours indicate the value of {$H_s=\sqrt{H_x^2+H_y^2}$} (blue=low, yellow=high).}
\label{fig:height_period_direction}
\end{figure}

\autoref{fig:height_period_direction} shows a scatter plot of $T_m$ against $H_x$ and $H_y$. (A plot of the normalised variables $(X_3,X_4,X_5)$ would look similar, but the non-normalised variables are shown here to aid physical interpretation). The conical shaped bound imposed by the limiting wave steepness is evident. Another feature that is apparent is that the data cloud is hollow on the side $H_x<0$. This is because of fetch limitations in this direction, meaning that waves propagating towards the west are steeper wind-driven waves, so that $H_s$ and $T_m$ are strongly correlated in this region. This violates the assumption of the distribution being star-shaped with respect to the choice of origin. However, the `edge region' that is not modelled corresponds to low values of $H_s$ at a given $T_m$ and wave direction, which is less critical for extreme responses. For the present choice of origin at $(H_x,H_y,T_m)=(0,0,6.2)$, it appears that there are some sharp changes in the distribution with the vertical angle, for rays towards the negative x direction. This might be hard to capture with an ANN. Nevertheless, we can assess how well the model performs, given these challenges.

\subsection{Assessment of angular density model} \label{sec:angular_assess}
\begin{figure}[ht]
\centering
\includegraphics[scale=0.65]{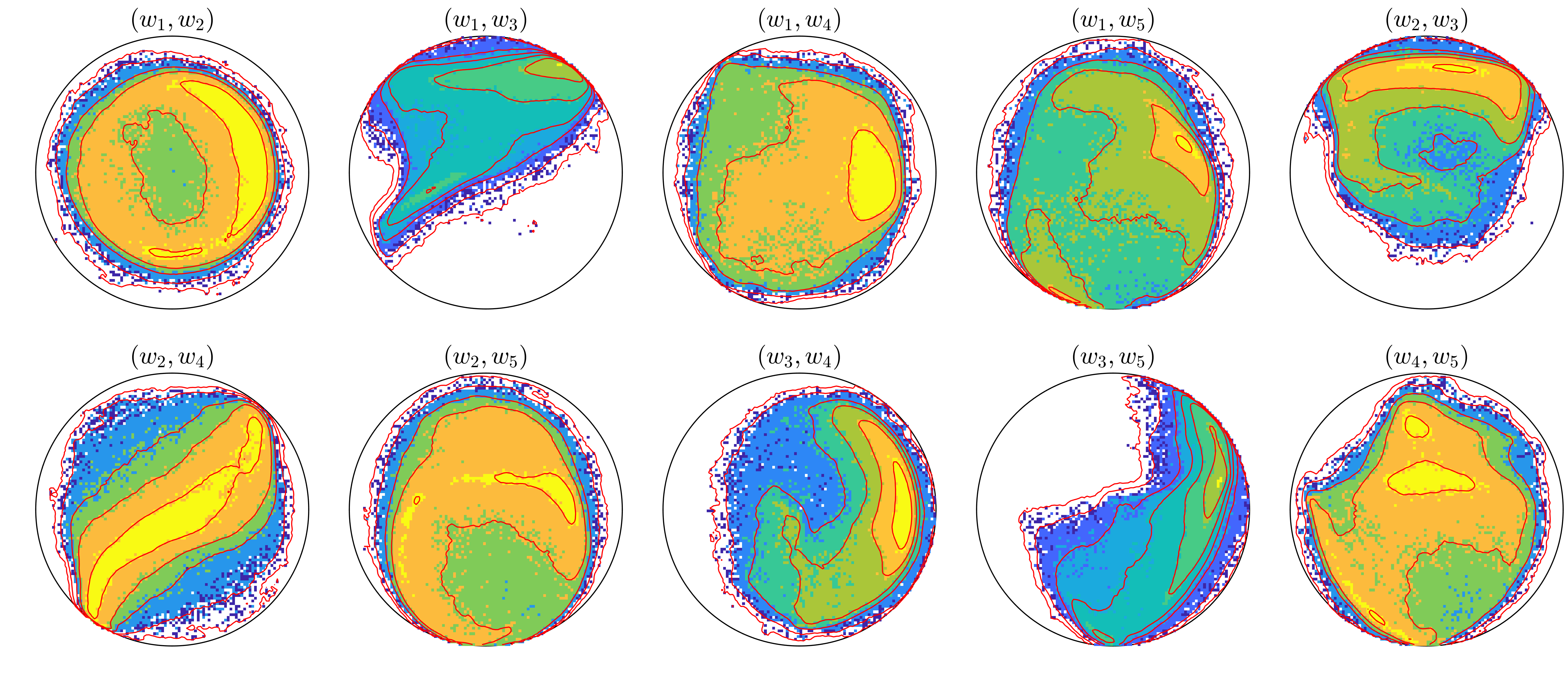}
\includegraphics[scale=0.65]{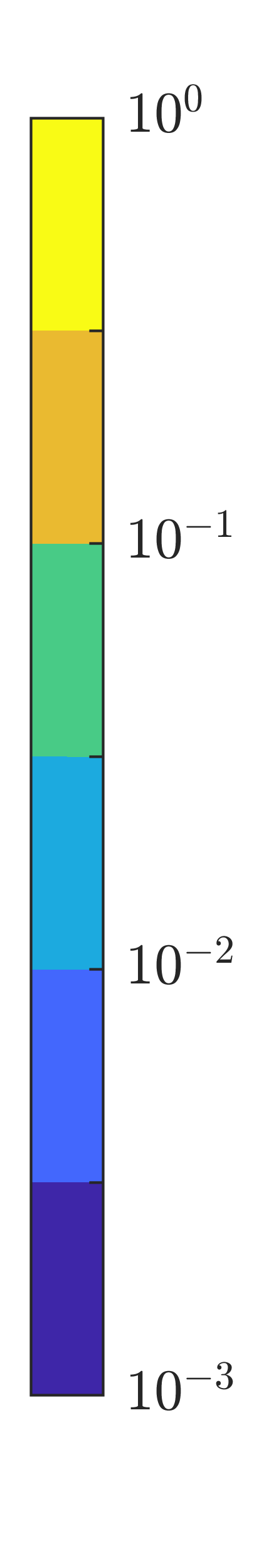}
\caption{Comparison of empirical joint densities of pairs of angular components (coloured plots) with contours of the joint densities from a sample from the KD model (red lines). Note that the colour resolution has been restricted so that the boundaries of the coloured regions correspond to the contour levels from the KD model.}
\label{fig:KD_contour}
\end{figure}

The assessments of the models for both the angular and conditional radial distributions are derived from a sample simulated from the fitted model, of 100 times the size of the original sample (i.e., $2.71\times 10^7$ samples, or approx. 3000 years of hourly data). The larger sample size reduces the uncertainty in estimates in the lower-probability regions, compared to the observed sample, and allows us to assess how well the model extrapolates from the observations. \autoref{fig:KD_contour} shows a comparison of the empirical joint densities of paired angular components $(w_i,w_j)$, $i,j\in\{1,...,5\}$, with contours of the corresponding joint densities from the KD model. For the empirical density of the observations, the colour resolution has been restricted so that the boundaries of the coloured regions correspond to the contour levels from the KD model. That is, contours from the KD model have been computed at density levels $10^{-3}, 10^{-2.5},...,10^{-0.5}$, and for the observations, all bins where the estimated density is in the interval, say $[10^{-2.5}, 10^{-2})$, have been assigned the same colour. Overall, the KD model compares very well with the observed densities, with the contours following the complex patterns in the observations. The smoothing effect of the model on the observations is evident, especially at low density levels.

\begin{figure}[ht]
\centering
\includegraphics[scale=0.65]{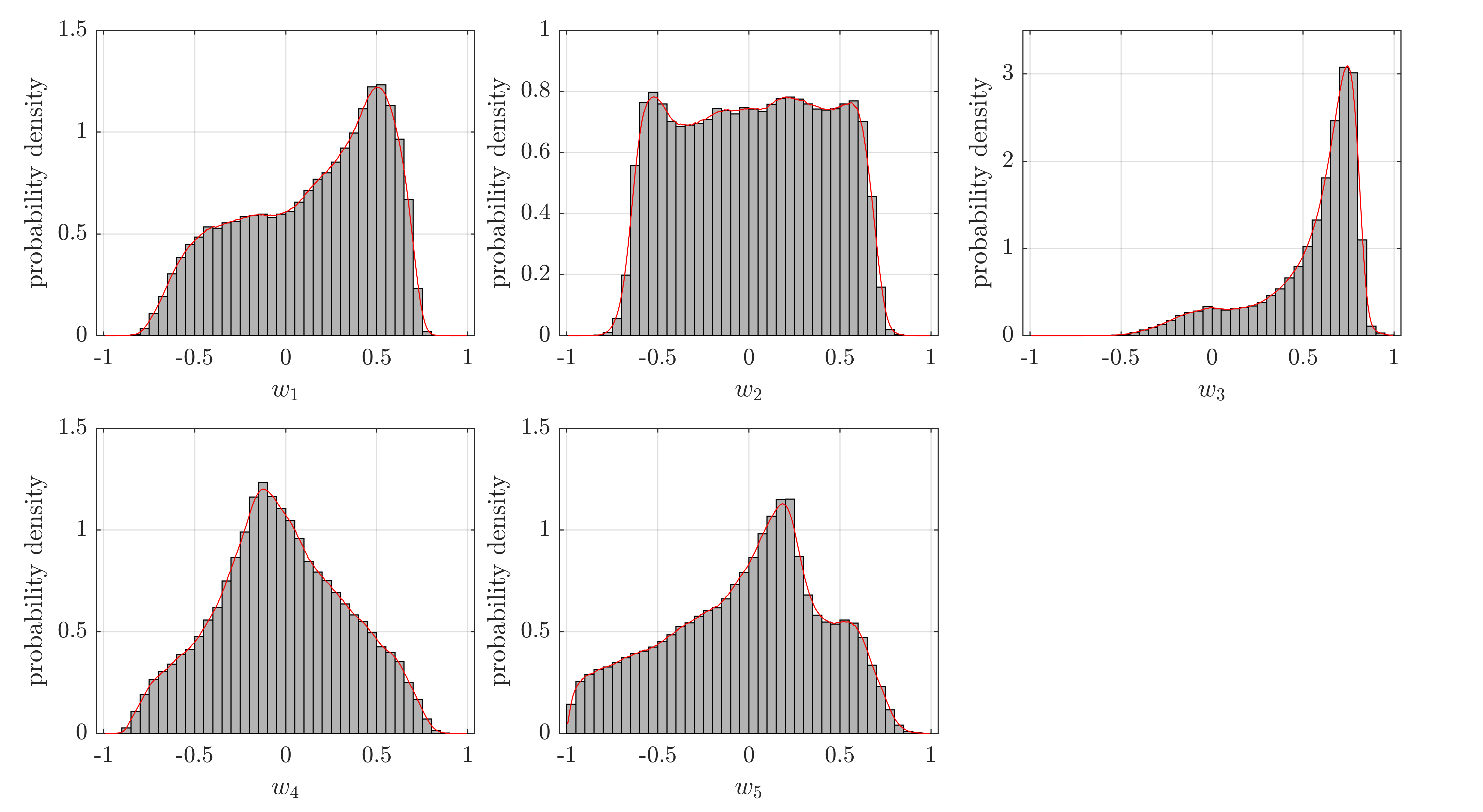}
\caption{Comparison of empirical marginal densities of angular components (histograms) with marginal densities from the KD model (red lines).}
\label{fig:KD_histogram}
\end{figure}

\begin{figure}[ht]
\centering
\includegraphics[scale=0.65]{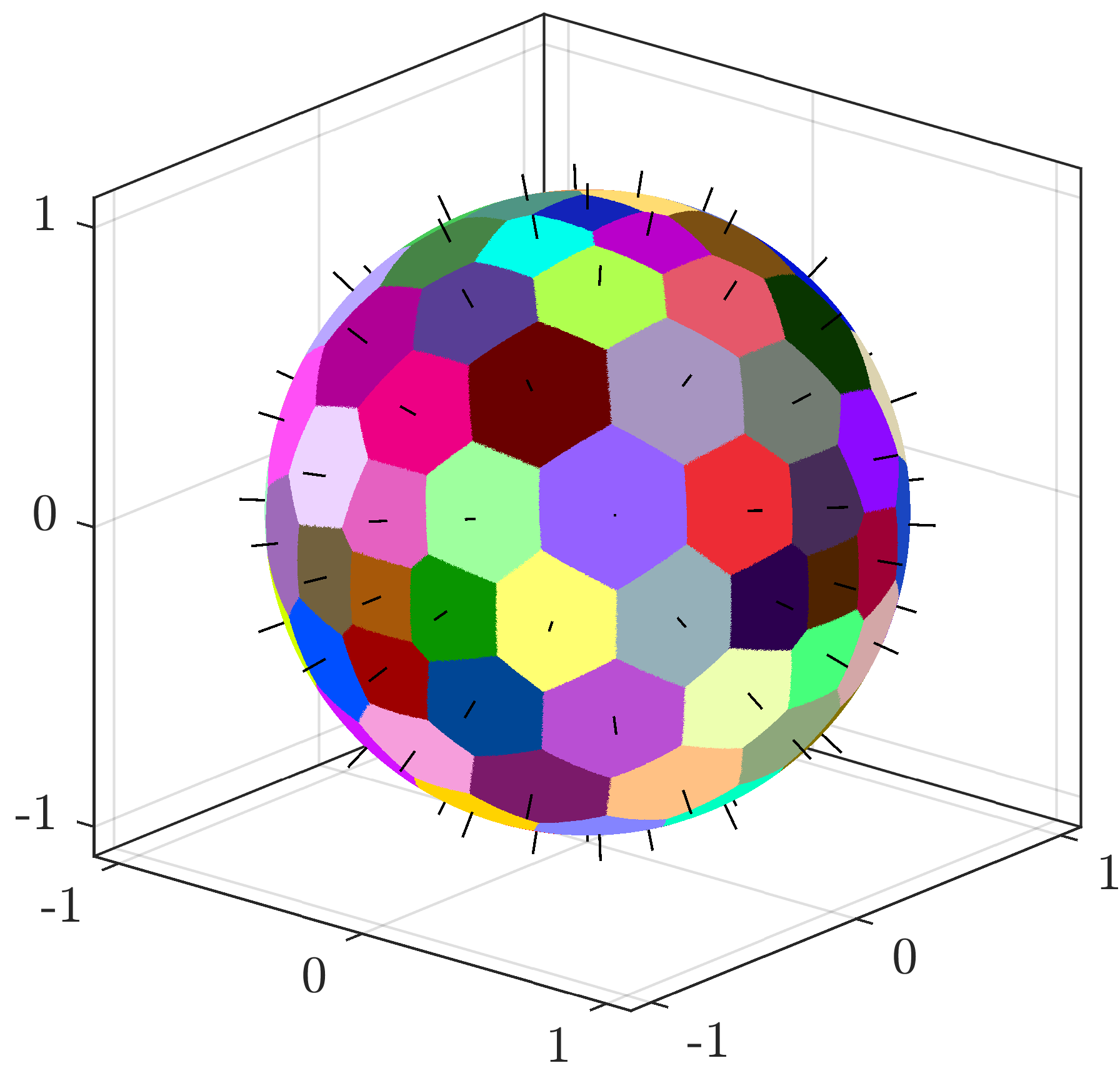}
\caption{Example of Voronoi partitioning of the sphere in $\RR^3$ into non-overlapping cells, corresponding to the set of points closest to each direction vector (black lines).}
\label{fig:angle_groups}
\end{figure}

Whilst \autoref{fig:KD_contour} shows that the KD model provides a good representation of dependence between pairs of angles, it does not consider how the model performs locally on small regions of $\mathbb{S}^4$. To motivate how this can be assessed in five dimensions, first consider the one-dimensional case. \autoref{fig:KD_histogram} shows the empirical marginal densities of each angular component depicted as histograms, compared with the marginal estimates from the KD model. When we assess the fit of the model in this way, we are making a visual assessment of the whether the number of observations in each bin agrees with the number predicted by the model. This idea can readily be extended to higher dimensions, by partitioning the space into discrete regions and comparing the number of observations in a region, with the expected number in that region from the model. In this case, we use a Voronoi partition relative to a set of pseudo-regularly spaced direction vectors, generated using the method described in Appendix \ref{app:L1L2_projection}. A Voronoi partition is a partitioning of a space, where each partition is the set of points closest to a given reference point. \autoref{fig:angle_groups} shows an example partitioning of the sphere in $\RR^3$ relative to a set of 102 pseudo-regularly spaced direction vectors. 

Suppose that we have a partitioning of $\dsphere$ into non-overlapping cells, together with a model from which we can calculate the probability, $p_i$, that an observation chosen at random falls into cell $i$. If the observations are independent, the number of observations, $N_i$, falling into cell $i$ follows a binomial distribution with CDF
\begin{equation} \label{eq:binom_CDF}
    \Pr(N_i\le k) = \sum_{j=0}^k \binom{n}{j}p_i^j(1-p_i)^{n-j}, \quad k\in\mathbb{N},
\end{equation}
where $n$ is the number of independent observations. In our application, the data are serially correlated, and hence violate the above assumption of independence. To mitigate for this, we down-sample the data to take one observation per day, resulting in a sample size of $n=11321$ independent observations. 

\begin{figure}[ht]
\centering
\includegraphics[scale=0.65]{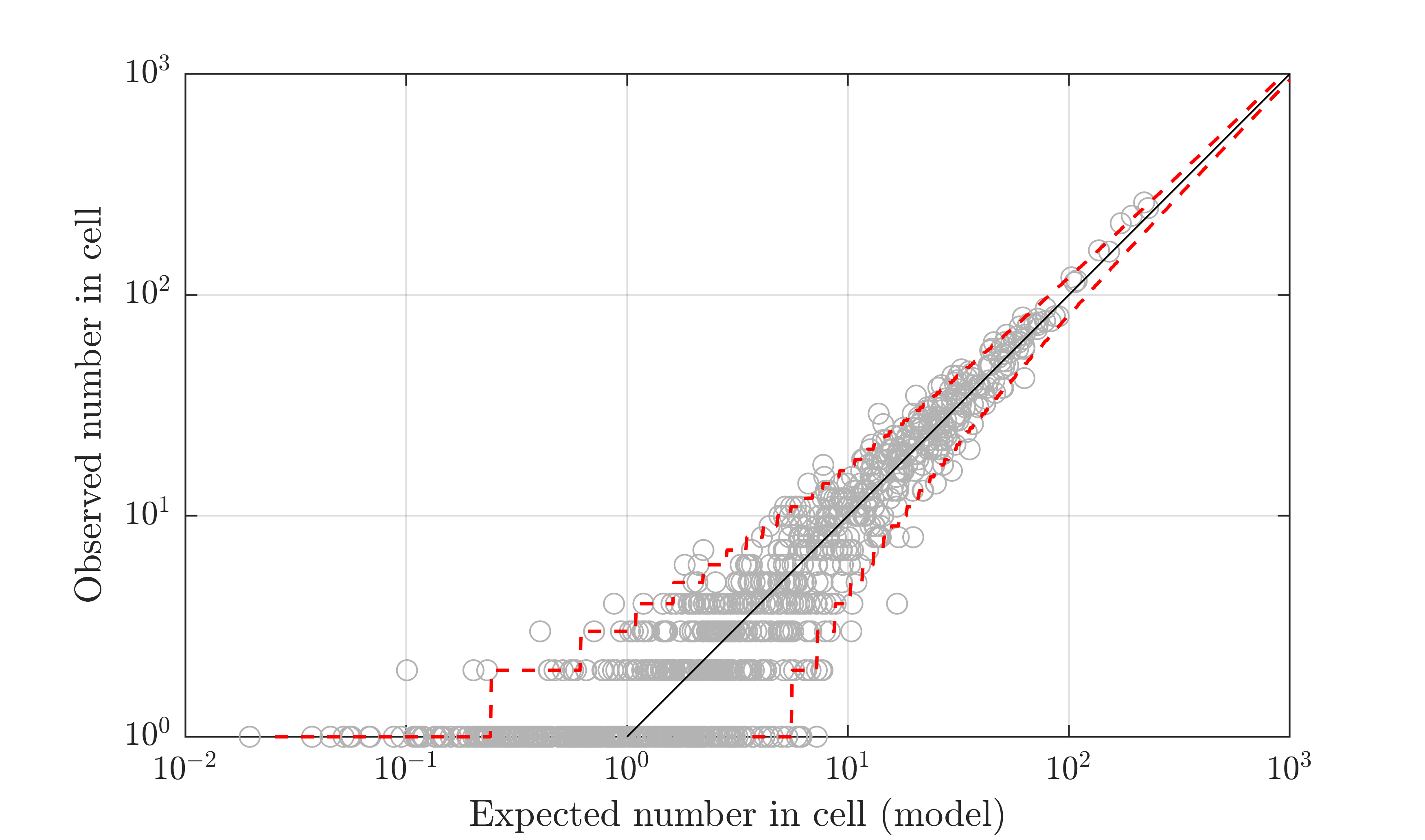}
\caption{Comparison of observed number of data points in a local angular cell against the expected number in the cell from the KD model, based on a Voronoi partition of $\mathbb{S}^4$. Solid black line shows 1:1 relation, red dashed lines show 95\% confidence interval on observed number based on sampling variability.}
\label{fig:KD_scatter}
\end{figure}

Equation \eqref{eq:binom_CDF} can be used to add confidence bounds to scatter plots of the observed number in a cell against the expected number predicted by the model. \autoref{fig:KD_scatter} shows a scatter plot of the number of observations in each cell of a Voronoi partition of $\mathbb{S}^4$ against the expected number from the KD model. The expected number in each cell is calculated as $\mathrm{E}(N_i) = n p_i$, where $n$ is the observed (down-sampled) sample size and $p_i$ is the fraction of the simulated sample falling into cell $i$. The Voronoi partition is based on a set of 5890 direction vectors, generated using $m=8$ in the method described in Appendix \ref{app:L1L2_projection}. However, only 1001 of the resulting cells contained observations, due to the non-uniformity of the angular distribution, discussed above. Overall, the model does very well at replicating the local characteristics of the density, both in the high and low density regions, with differences attributable to sampling variability. 

\subsection{Assessment of conditional radial model}
\label{sec:radial_assess}
\begin{figure}[ht]
\centering
\includegraphics[scale=0.7]{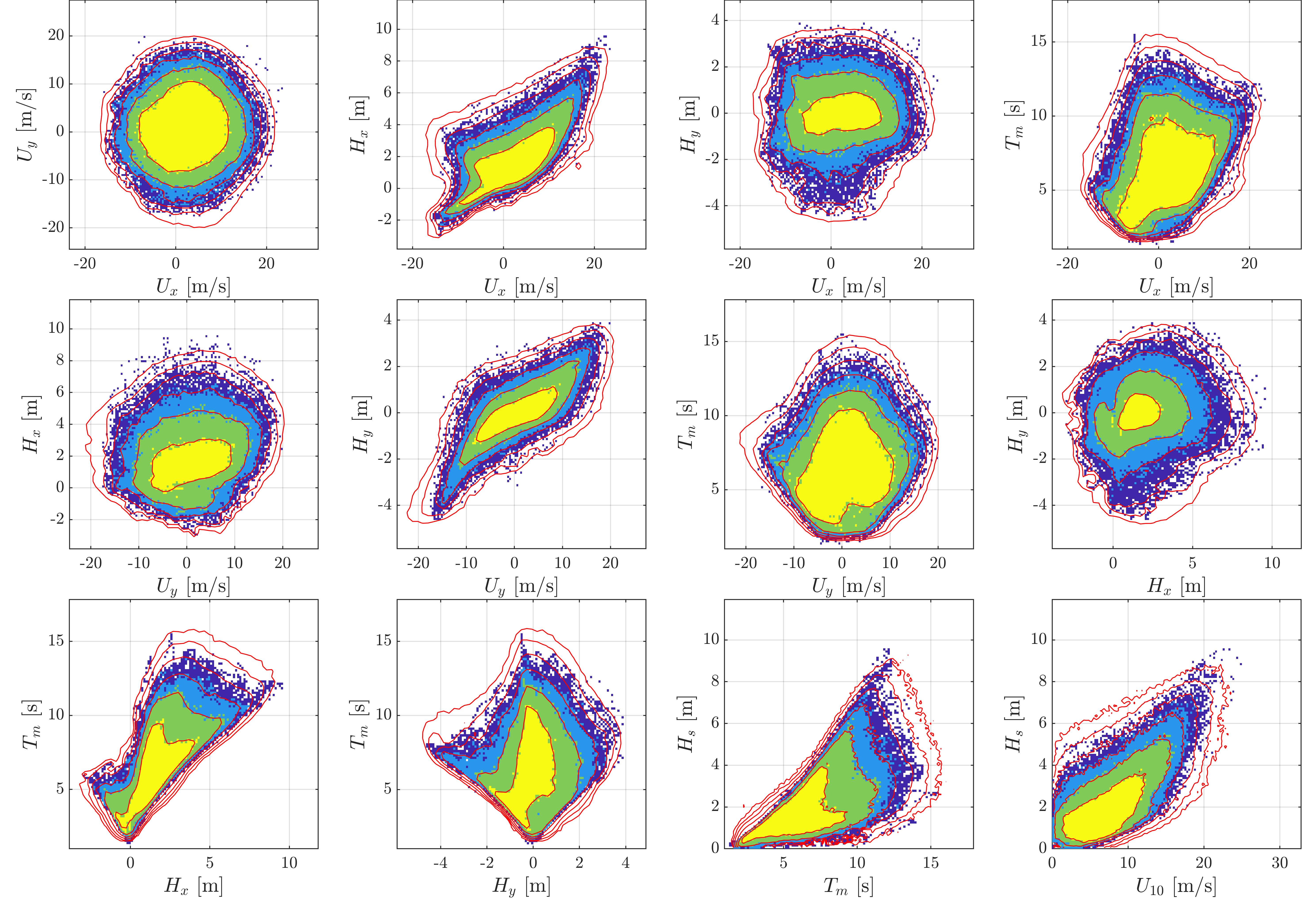}
\caption{Comparison of empirical joint densities of pairs of variables (coloured plots) with contours of the joint densities from a sample from the SPAR model of 100 times the original sample size (red lines). Note that the colour resolution has been restricted so that the boundaries of the coloured regions correspond to the contour levels from the SPAR model. Contours at equal logarithmic increments. (See Section \ref{sec:data} for nomenclature). }
\label{fig:pairwise_contour}
\end{figure}

We assess the quality of the conditional radial model in several ways, analogous to the diagnostics for the angular model. The ability of the model to recreate the pairwise relations is considered in \autoref{fig:pairwise_contour}, which shows empirical joint densities of pairs of the observed variables (on the orignal scale) compared with contours of the joint density from the fitted model. Although the SPAR model gives an explicit representation of the joint density, this cannot be projected into lower dimensions without computationally-expensive integration. Instead, the joint densities from the model are estimated empirically from a simulated sample from the model. As such, the density shown is subject to sampling uncertainty in the low density regions, but less than that for the observations, due to the larger simulated sample size. Within the body of the data (i.e., below the threshold level) the observations are resampled rather than modelled. Good agreement between the model contours and observations is therefore to be expected in the bodies of the data clouds. However, the density contours in the outer `extreme' regions correspond to points which are simulated from the non-stationary GP model. Overall, the simulated relationships between variables closely follow those of the observations. The model predicts some larger values of steep waves than observed. However, these occur at lower wave heights, so are less critical for design purposes. 

\begin{figure*}[ht]
\centering
\includegraphics[scale=0.8]{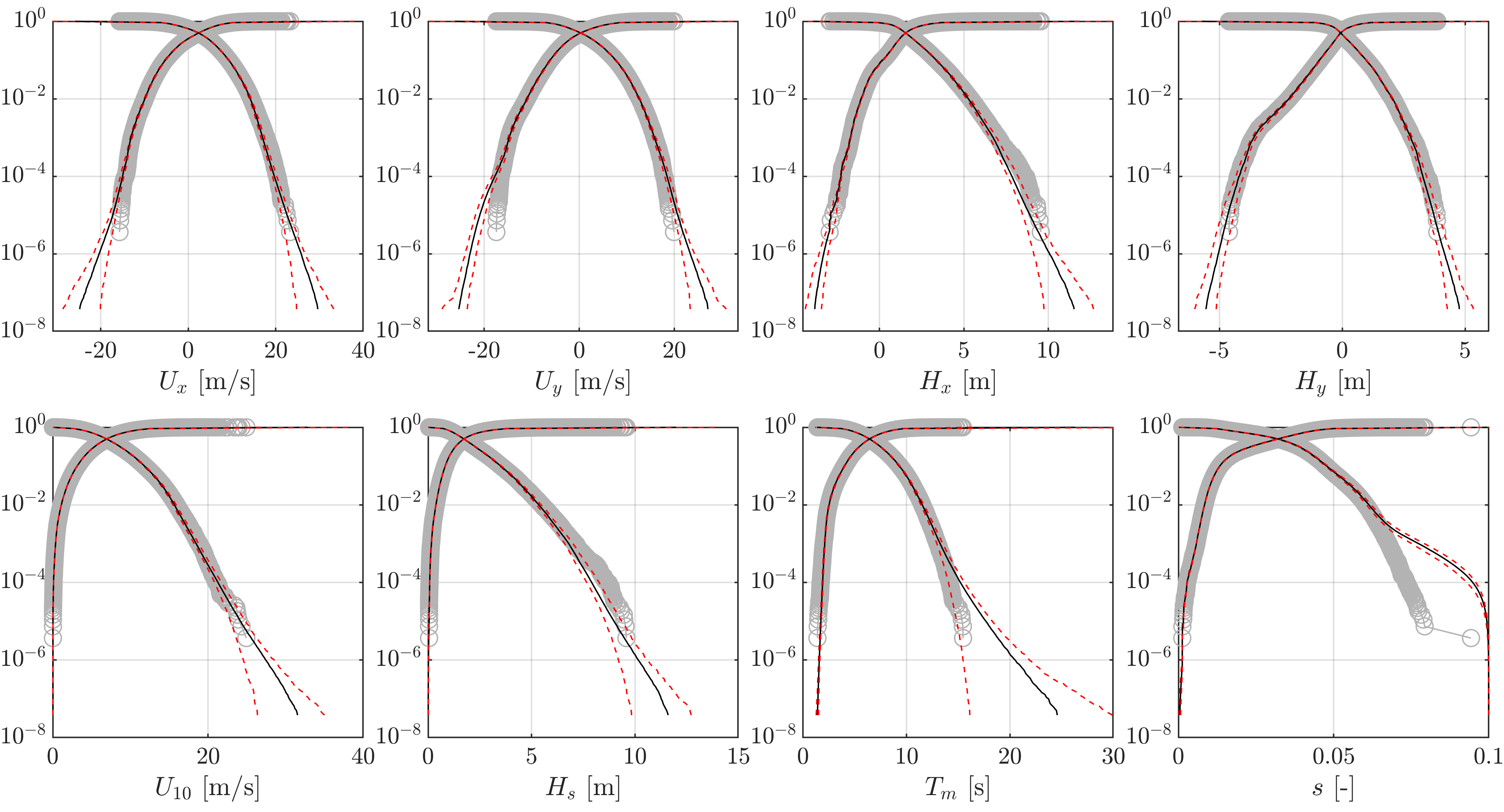}
\caption{Exceedance and non-exceedance probabilities of marginal and related variables for observed (grey circles) and simulated (black lines) data. The red dashed lines indicate a 95\% confidence interval for the model predictions based on 200 bootstrap trials. (See Section \ref{sec:data} for nomenclature).}
\label{fig:marginal_tails}
\end{figure*}

As the SPAR model does not model the tails of the marginal variable directly, it is also useful to assess how well the simulated data matches the observed marginal tails. \autoref{fig:marginal_tails} shows exceedance and non-exceedance probabilities on a logarithmic scale for the five variables used in the model, as well as the derived variables $H_s$, $U_{10},$ and $s$. The model predicted some unrealistically steep conditions, so we have removed data points where steepness exceeded a value of 0.1. This corresponded to approximately 0.01\% of the simulated sample and small wave heights ($H_s<3$~m). With these points removed, overall, the model performs well and provides a good match for both the upper and lower observed tails. There is a slight tendency to underestimate the upper tail of $H_x$, which results in a slight underestimation of the upper tail of $H_s$.

To assess the uncertainty in the fitted model, a block bootstrap method \cite{Kunsch1989} has been used. The block bootstrap is used to account for the serial correlation in the observations. A block size of four days was used in this case, corresponding to typical storm durations at this location. The data were resampled 200 times, and the model was fitted to each resampled dataset. For each fitted model, a sample of 100 times the size of the original dataset was generated, and used to construct an empirical 95\% confidence interval for the marginal quantiles, shown in \autoref{fig:marginal_tails}. In general, the level of uncertainty for the wave height and wind speed variables is low. However, the uncertainty bounds on the upper tail of the mean wave period, $T_m$, are rather wide, and indicate unrealistically large values from some bootstrap trials. 

\begin{figure}[ht]
\centering
\includegraphics[scale=0.8]{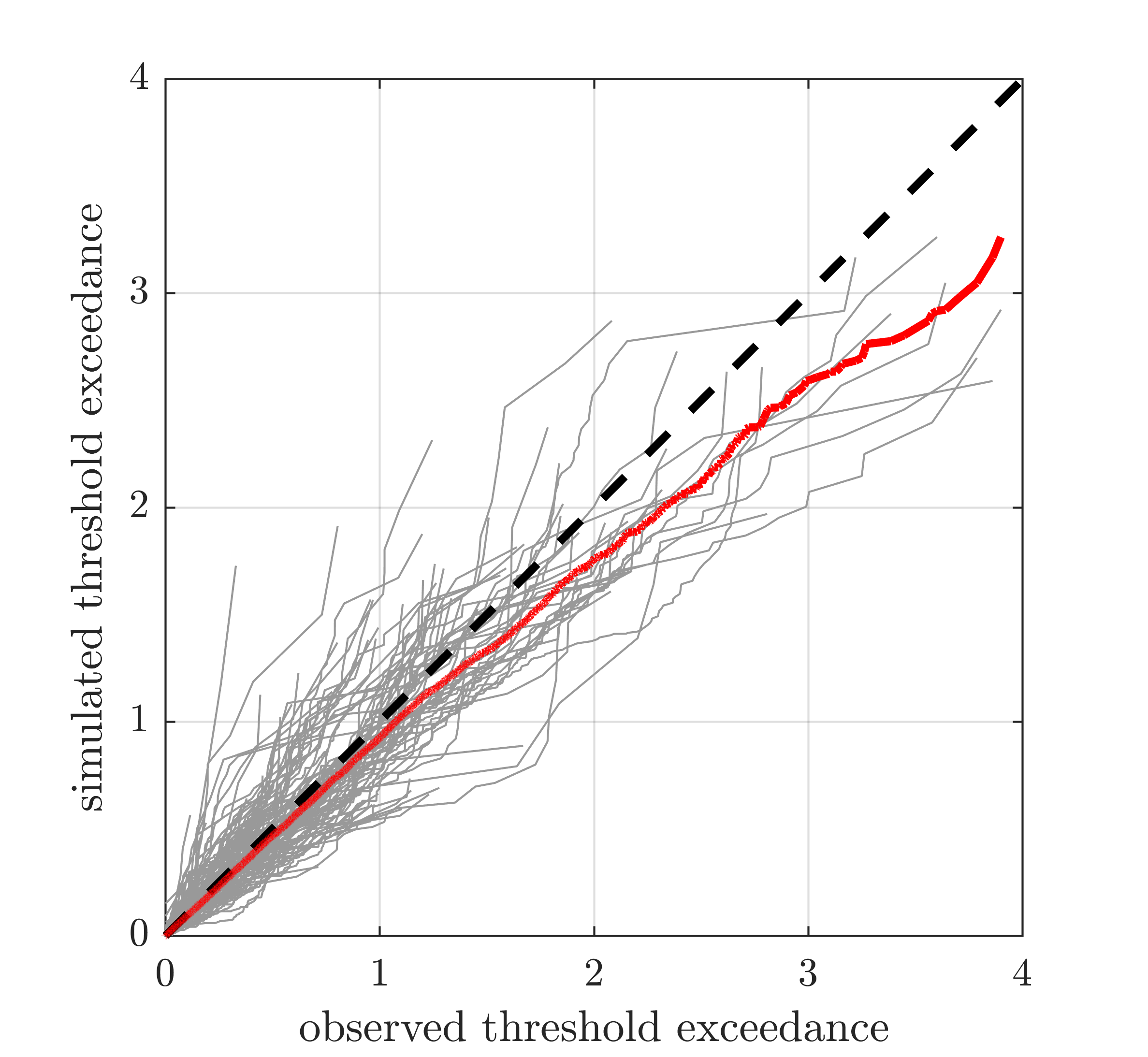}
\caption{QQ plots of simulated versus observed threshold exceedances in bins of $15^\circ$ radius from a grid of points on $\mathbb{S}^4$. Grey lines are individual bins; red line is aggregated value over all bins; black dashed line is equality.}
\label{fig:local_tails}
\end{figure}

Finally, to give an indication of the variation in model performance over the angular domain, \autoref{fig:local_tails} shows QQ plots of simulated against observed threshold exceedances, binned over small angular ranges. As above, we have used a grid of 1002 fixed angles and a radius of $15^\circ$ to define the angular bins. Only bins with 200 or more observations have been used, so that there are approximately 20 or more threshold exceedances per bin. To account for the variation of the threshold level over the bins, the threshold is taken as the empirical quantile at exceedance level $\zeta$ from the simulated dataset (this approximates an average of the non-stationary threshold over the bin). The simulated data are then interpolated to the same probability levels as the observations. There is some scatter between the observed and simulated values. However, the aggregated trend shows good agreement, although with a slight tendency to underestimate at larger values. Although the total sample size is relatively large, the number of observations in local angular regions can be quite small. The underestimation from the model may therefore be related to the use of maximum likelihood estimation, which is known to produce a slight negative bias in quantile estimates for small sample sizes \cite{hosking1987parameter}. 

\section{Discussion}
\label{sec:discussion}
The use of artificial neural networks for modelling the angular variation of the radial threshold GP parameters was motivated by the computational concerns discussed in Section \ref{sec:rad_dens}. We note that selecting an architecture for any neural network is non-trivial, and care must be taken to ensure the resulting model offers sufficient flexibility without overfitting. We also remark that, in general, neural networks require a large amount of data for accurate model fitting \cite{goodfellow2016deep}; this is generally a challenge for modelling extremes since, by definition, very little data are available. It is currently not clear what sample sizes, tuning parameters, or architectures are required to accurately fit the SPAR model via deep learning, and this should be explored in further work. In the present approach, we have used 80\% of the data for training and 20\% for validation. The validation data are used to avoid overfitting. It is possible that using only 20\% of observations for validation is insufficient to force a sufficiently smooth solution in regions of sparse angular observations. An alternative is to use a full cross-validation scheme, in which the model results are averaged over, e.g., five fits using a different 20\% of the data for testing. This would make better use of the limited observations, although with increased computational cost. Averaging over multiple plausible candidate models provides a general means of accommodating uncertainty due to modelling choices that are somewhat arbitrary, like the location of the polar origin, the threshold level, and neural network architecture. 

One notable observation from Section~\ref{sec:application} was that the choice of origin for defining the SPAR model was non-trivial. Initially, we naively assumed that the componentwise mean was a suitable origin, but this resulted in issues when fitting the model, as the data cloud was not star-shaped with respect to this initial choice of origin. Selecting an appropriate origin for the SPAR model beyond the lower dimensional ($d \leq 3$) setting remains a challenge as full data visualisation is not possible, and inappropriate choices can invalidate the modelling assumptions. Future work could explore the robust selection of the origin in a data-driven manner, removing the need for domain-specific knowledge. 

In the current work, we have included the wind and wave directions in the model by considering the x- and y-components of the wind speed and wave height. Another way to model the five considered variables is to treat the wind and wave directions as covariates, and then model the joint distribution of wind speed, wave height, and wave period conditional on these two covariates. Initial investigations suggest that this could resolve some of the issues in the current approach caused by the complex shape of the data cloud. If the origin of the polar coordinate system is allowed to vary with wind and wave direction, then the remaining three-dimensional data clouds appear more star-shaped, potentially resulting in a better fit for the model. This will be investigated in future work. Note that if we were to model a two-dimensional case of, say, wave height and direction, using the SPAR approach, then this is identical to fitting a non-stationary model to $H_s$ with wave direction as a covariate. Indeed, this was partially the motivation for the SPAR approach. However, when more variables are involved, treating directions as covariates or considering the joint distribution of x- and y-components does represent different sets of modelling assumptions. 

From Section~\ref{sec:application}, it was also clear that as the dimension of the data increases, so too does the sparsity. This is particularly apparent in \autoref{fig:ang_dens}, where we observed some orthants containing just a handful of observations. Inference within such regions is problematic, since there is very little data to train the neural network model. We also noted that in sparse regions, the resulting parameter estimates did not always respect the physical features of the data, e.g., upper bounds of the variables. Such issues are not unique to the SPAR model, and one would expect to encounter the same problems with alternative modelling approaches when applied in high dimensions. The effect of dimensionality on data sparsity is not necessarily intuitive to understand. For example, generating pseudo-regularly spaced points on $\mathbb{S}^4$ in the manner described in Appendix \ref{app:L1L2_projection}, with an average spacing of approximately $4.5^\circ$ (corresponding to $m=20$ points along the positive half of each dimension), results in 216,002 direction vectors. This corresponds to $216002/(24\times 365.25) \approx 24.6$ years of hourly observations -- a similar order of magnitude to the size of dataset considered here (although, as discussed above, serial correlation in the data reduces the effective sample size). Compare this to the spacing of the same number of direction vectors in two dimensions, which would be $360/216022 \approx 0.0017^\circ$. So, a dataset that gives a high directional coverage in two dimensions gives a much more sparse directional coverage in five dimensions. Moreover, the sparsity increases exponentially with the number of dimensions, meaning that this type of analysis (or any multivariate density modelling method) will require an exponential increase in dataset length to maintain a similar level of mean spacing between observed angles. Future work could explore whether notions of sparsity can be incorporated into the SPAR framework to improve the robustness and efficiency of the model. 

In the present implementation, the model is only estimated for observations above the threshold level, with the threshold exceedance probability, $\zeta$, chosen by testing models fitted with different values of $\zeta$. A possible extension would be to model the entire radial distribution with a blended model that takes a semi-parametric form in the bulk of the distribution and a parametric GP form in the tail, similar to \cite{hansen2020directional, majumder2025semi}. This has the advantage that the uncertainty in threshold estimation is incorporated in the inference. 

Finally, we note that although we have accounted for serial correlation in the data in some of the model fitting and diagnostics, the resulting SPAR model is a model for all observations. As such, when we simulate from the model, we do not account for the serial correlation. This could potentially be addressed by using a two-step simulation procedure, where we first simulate a point from the model, then simulate a section of time series which passes through the simulated point. The time series simulation could be accomplished using a resampling approach \cite{mackay2020estimation} or by modelling the temporal evolution of variables conditional on a peak value in a similar manner to \cite{tendijck2023temporal}. This will be investigated in future work. 

\section{Conclusions}
\label{sec:conclusions}
In this work, we have introduced a novel deep learning framework for inference with the SPAR model. The computational scalability and robustness of neural networks result in a modelling approach which requires very few assumptions, offers a high degree of flexibility, and can be applied in higher-dimensional settings compared to existing techniques. We use our approach to approximate the complex joint tail behaviour of a five-dimensional metocean dataset, with diagnostics indicating our model is able to accurately represent the observed dependence structure. Given the complex dependence observed in the data, the MLP model for the angular variation of the GP parameters performs very well in representing the `extremes' of the dataset. By `extremes', we are referring not just to the largest and smallest values of each observed variable, but anywhere on the outer part of the data cloud. Moreover, the SPAR model provides an asymptotically justified basis for extrapolating outside the range of observations. Simulation from the fitted SPAR model subsequently allows ones to generate large, physically-realistic event sets, allowing practitioners to easily perform robust risk assessments and estimate probabilities of structural failure.

\section*{Acknowledgment}
EM was funded by the EPSRC Supergen Offshore Renewable Energy Hub, United Kingdom [grant no: EP/Y016297/1]. This work has made use of the resources provided by the Edinburgh Compute and Data Facility (ECDF) (\url{http://www.ecdf.ed.ac.uk/})

\begin{appendices}

\section{Pseudo-regularly spaced points on the hypersphere} \label{app:L1L2_projection}
\begin{figure}[ht]
\centering
\includegraphics[scale=0.6]{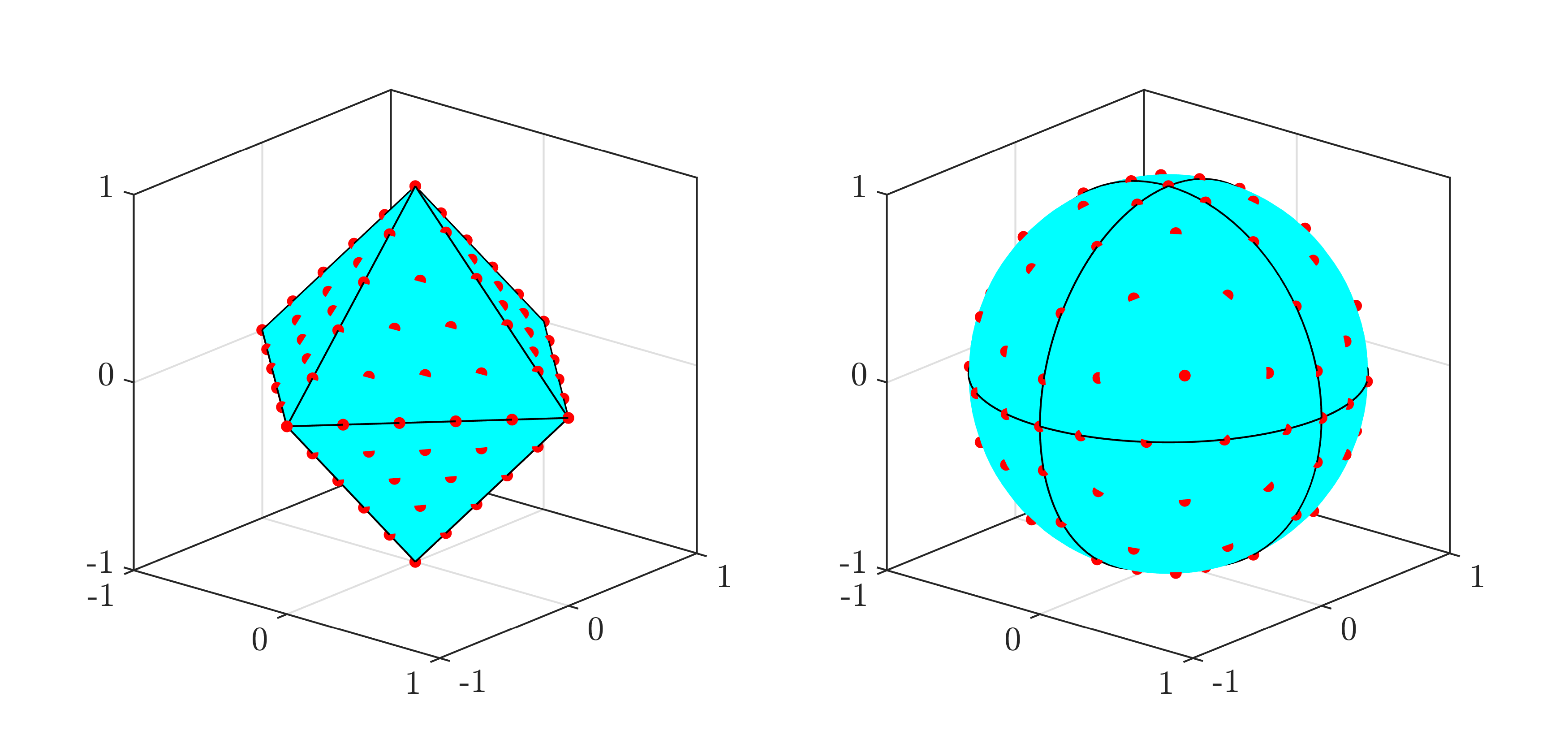}
\caption{Illustration of the mapping of regularly spaced vectors on the surface of the $L^1$ unit sphere (left) onto the $L^2$ unit sphere (right).}
\label{fig:dir_vectors}
\end{figure}

To generate a grid of pseudo-regularly spaced points on the hypersphere, we take the approach proposed in \cite{mackay2023-DIFORM}, and define a regular grid of points on the $L^1$ sphere and project this onto the surface of the $L^2$ sphere. This is computed by creating a regular grid of points in the cube $[-1,1]^d$, with $2m+1$ points along each dimension: $U_{grid}=\{(i_1/m, ..., i_d/m) : i_j\in\{-m,...,m\}, j=1,...,d\}$. Then, we define $U_1 = \{ \mathbf{u} \in U_{grid} : \|\mathbf{u}\|_1=1\}$, where $\|\cdot\|_1$ is the $L^1$ norm, given by $\|\mathbf{u}\|_1 = |u_1| + \cdots + |u_d|$ for $\mathbf{u} = (u_1,...,u_d)$. Finally, we compute a set of direction vectors $U\subset \mathbb{S}^{d-1}$ by $U = \{ \mathbf{u} / \|\mathbf{u}\|_2 : \mathbf{u}\in U_1 \}$. This is illustrated in \autoref{fig:dir_vectors} for the case $m=5$ and $d=3$. 

\section{Simulation from the power spherical distribution} \label{app:PSD_simulation}
\begin{figure}
	\centering
	\includegraphics[scale=0.25]{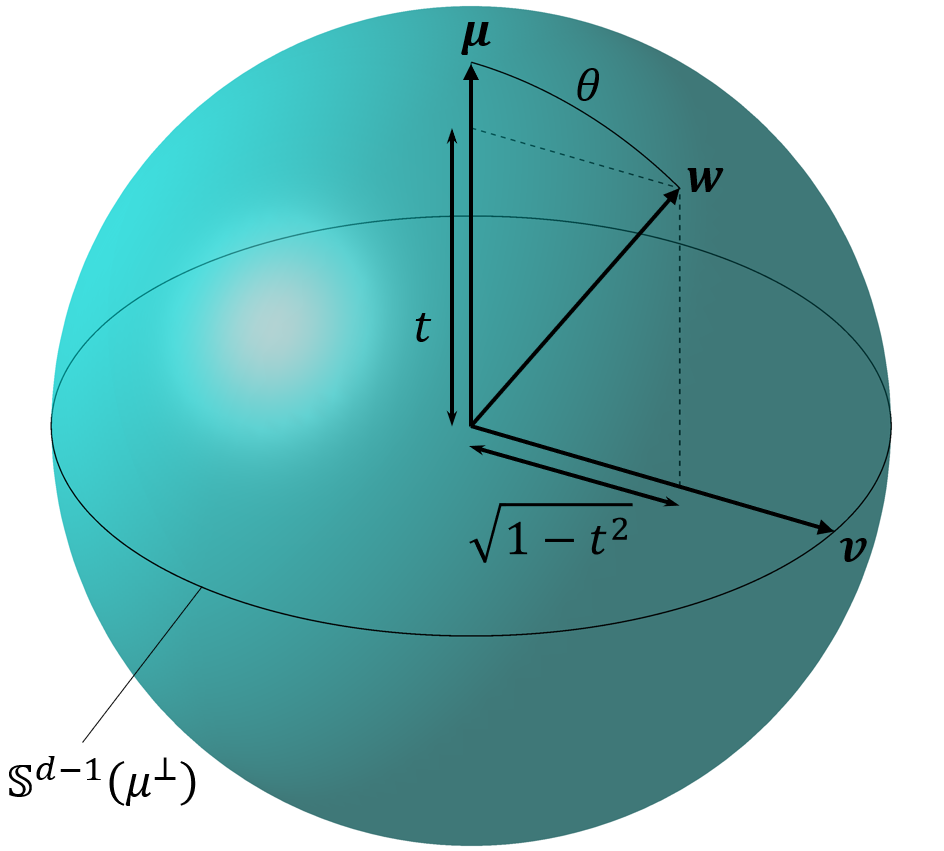}
	\caption{Illustration of tangent-normal decomposition of a vector $\mathbf{w}\in\mathbb{S}^2$ relative to $\bm{\mu}\in\mathbb{S}^2$.}
	\label{fig:tangent_normal}
\end{figure}

A method for simulating from the PS distribution was presented in \cite[][Algorithm 1]{deCao2020psd}. Here we present some background details on simulating from rotationally-symmetric distributions on $\dsphere$, to illustrate why the PS distribution offers computational advantages over the more commonly-used von Mises-Fisher (vMF) distribution. Further details can be found in standard texts on directional statistics, e.g., \cite{mardia2000, ley2017modern}. An essential tool for modelling random vectors on the sphere is the \textit{tangent-normal decomposition}, illustrated in \autoref{fig:tangent_normal}. For any vectors $\mathbf{w},\bm{\mu}\in\dsphere$ we can express $\mathbf{w}$ in terms of components that are aligned to $\bm{\mu}$ and tangential to it:
\begin{align*}
	\mathbf{w} &= t \bm{\mu} + \sqrt{1-t^2} \mathbf{v},\\
	&= \cos(\theta) \bm{\mu} + \sin(\theta) \mathbf{v},
\end{align*} 
where $t=\cos(\theta) = \mathbf{w}^\top \bm{\mu}$ and $\mathbf{v}=(\mathbf{w}-t\bm{\mu})/\|\mathbf{w}-t\bm{\mu}\|_2$. The unit vector $\mathbf{v}$ lies in the tangent space to $\bm \mu$, $\dsphere(\bm{\mu}^\perp) \coloneq \{\bm{x}\in \dsphere : \bm{x}^\top \bm{\mu} = 0\}$.  

A density $f:\dsphere\to [0,\infty)$ that is rotationally symmetric about its location $\bm{\mu}\in\dsphere$ can be expressed in terms of an angular function $f_a : [-1,1] \to [0,\infty)$, as $f(\mathbf{w}) = f_a( \mathbf{w}^\top \bm{\mu})$ for all $\mathbf{w}\in\dsphere$. That is, the density at $\mathbf{w}$ just depends on the distance from $\mathbf{w}$ to $\bm\mu$ along the surface of the sphere. The tangent-normal decomposition of a rotationally symmetric random vector leads to the following useful result \cite{watson1983}. Suppose that random vector $\mathbf{W}\in \dsphere$ has density $f$ that is rotationally symmetric about $\bm{\mu}$, with corresponding angular function $f_a$. Define the tangent and normal components of $\mathbf{W}$ as above, with $T = \mathbf{W}^\top \bm{\mu}$ and $\mathbf{V} = (\mathbf{W}-T\bm{\mu})/\|\mathbf{W}-T\bm{\mu}\|_2$. Then $T \in [-1,1]$ and $\mathbf{V}\in \dsphere(\bm{\mu}^\perp)$ are independent and $\mathbf{V}$ is uniformly distributed in $\dsphere(\bm{\mu}^\perp)$. Moreover, $T$ has density
\begin{equation} \label{eq:spherical_fT}
	f_T (t) = \frac{2 \pi^{(d-1)/2}}{\Gamma((d-1)/2)}  f_a(t) (1-t^2)^{(d-3)/2}.
\end{equation}
The distribution of $T$ is referred to as the marginal distribution of $\mathbf{W}$. In words, this result states that we can represent any rotationally symmetric distribution, in terms of one component, $T$, that describes the distance from the mean direction, and an orientation relative to the mean direction, $V$. Due to the rotational symmetry, the distance and orientation are independent.

The tangent-normal decomposition gives a computationally efficient method for simulating from rotationally symmetric distributions \cite{ulrich1984}. First note that if $\bm{e}_d=(0,...,0,1)^\top$, then $\dsphere(\bm{e}_d^\perp)$ is the embedding of $\mathbb{S}^{d-2}$ into $\RR^d$ (see \autoref{fig:tangent_normal} - in this example $\mathbb{S}^2(\bm{e}_3^\perp)$ is the embedding of the unit circle into $\RR^3$). Therefore, if we generate a random $T$ from the marginal distribution of $f$ and a uniform random $\mathbf{V}\in\mathbb{S}^{d-2}$, then $\bm{Y} = (T;\sqrt{1-T^2} \mathbf{V})$ has density $f$ with location vector $\bm{e}_1 = (1,0,...,0)^\top$. In then remains to rotate $\bm{Y}$ to the appropriate orientation $\bm{\mu}$. This can be achieved using a Householder transformation \cite{householder1958}. The Householder transformation describes a reflection about a (hyper)plane containing the origin with with unit normal vector $\bm{u}$. The transformation can be expressed in terms of the Householder matrix, defined as $\mathbf{P}=\mathbf{I}_d - \bm{u} \bm{u}^T$, where $\mathbf{I}_d$ is the identity matrix and $\bm{u} \bm{u}^\top$ is the outer product. For our application, the plane we wish to reflect about has normal vector $\bm{u} = (\bm{e}_1 - \bm{\mu})/\|\bm{e}_1 - \bm{\mu}\|_2$. Finally, a random vector with mean direction $\bm{\mu}$ is obtained as $\mathbf{W}=\mathbf{P}\bm{Y}$.

Simulation of uniformly distributed points on $\mathbb{S}^{d-2}$ is straightforward. If $N_1,...,N_{d-1}$ are independent standard normal variables, then $\mathbf{V}=(N_1,...,N_{d-1}) / \|(N_1,...,N_{d-1})\|_2$ is uniformly distributed on $\mathbb{S}^{d-2}$ \cite{muller1959note}. The key feature of the PS distribution that makes it efficient to simulate from, is that the marginal variable $T$ is defined in terms of an affine transformation of a Beta-distributed variable \cite{deCao2020psd}. Namely, $T=2Z-1$, where $Z\sim \text{Beta}(\alpha,\beta)$ with $\alpha=\kappa+(d-1)/2$ and $\beta=(d-1)/2$, where $\kappa$ is the bandwidth parameter (this follows from substituting the expression for the density of the PS distribution \eqref{eq:PSD_orig} into \eqref{eq:spherical_fT}). In contrast, the marginal distribution of the vMF distribution is not related to common distribution types, meaning that rejection sampling is necessary to sample from the marginal, which is computationally inefficient.

\end{appendices}
\printbibliography

\end{document}